\documentclass{article}

    \PassOptionsToPackage{numbers, compress}{natbib}

\usepackage[preprint]{neurips_2024}

\usepackage[utf8]{inputenc} 
\usepackage[T1]{fontenc}    
\usepackage{hyperref}       
\usepackage{url}            
\usepackage{booktabs}       
\usepackage{amsfonts}       
\usepackage{nicefrac}       
\usepackage{microtype}      
\usepackage[dvipsnames,svgnames]{xcolor}         

\title{HairFastGAN: Realistic and Robust Hair Transfer with a Fast Encoder-Based Approach}

\author{
  Maxim Nikolaev$^{1,3}$,~~Mikhail Kuznetsov$^{1, 2, 3}$,~~Dmitry Vetrov$^{4}$,~~Aibek Alanov$^{1, 3}$ \\
  $^1$HSE University,~~$^2$Skolkovo Institute of Science and Technology,~~$^3$AIRI \\
  \texttt{\{m.nikolaev, m.k.kuznetsov, alanov\}@2a2i.org}\\
  $^4$Constructor University, Bremen\\
  \texttt{dvetrov@constructor.university} \\
}

\usepackage{amsmath}
\usepackage{graphicx}
\usepackage{booktabs}
\usepackage[capitalize]{cleveref}

\usepackage{array}
\usepackage{multirow}
\usepackage{pifont}
\usepackage{makecell}
\usepackage{capt-of}
\begin{document}

\vspace{-0.5cm}
\maketitle
\vspace{-0.1cm}
\begin{center}
    \centering
    \vspace{-0.5cm}
    \includegraphics[width=\textwidth]{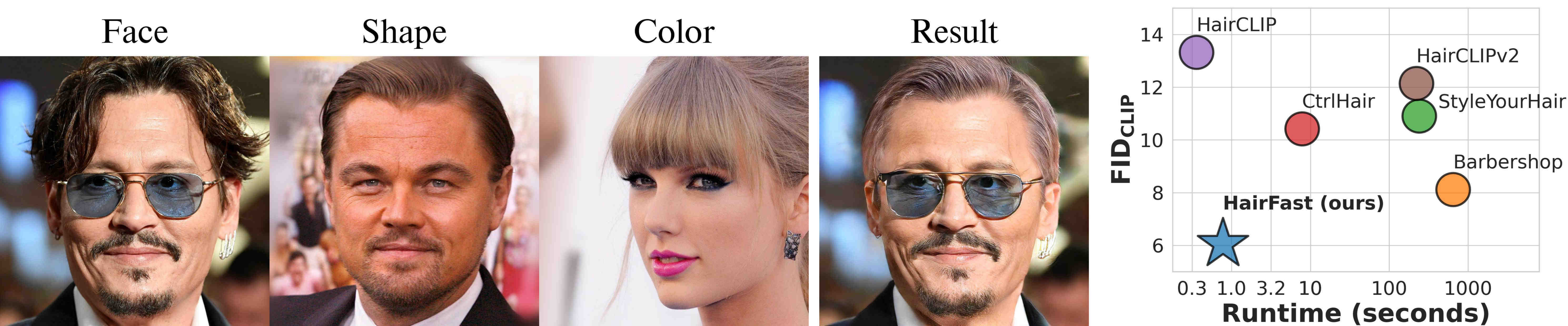}
    \vspace{-0.7cm}
    \captionof{figure}{{\bf HairFastGAN: Realistic and Robust Hair Transfer with a Fast Encoder-Based Approach.} {\it Our method takes as input a photo of the face, desired shape and hair color and then performs the transfer of the selected attributes. You can also see a comparison of our model with the others in the right plot. We were able to achieve excellent image realism while working in near real time.}
    }
\label{fig:teaser}
\end{center}

\begin{abstract}
Our paper addresses the complex task of transferring a hairstyle from a reference image to an input photo for virtual hair try-on. This task is challenging due to the need to adapt to various photo poses, the sensitivity of hairstyles, and the lack of objective metrics. The current state of the art hairstyle transfer methods use an optimization process for different parts of the approach, making them inexcusably slow. At the same time, faster encoder-based models are of very low quality because they either operate in StyleGAN's W+ space or use other low-dimensional image generators. Additionally, both approaches have a problem with hairstyle transfer when the source pose is very different from the target pose, because they either don't consider the pose at all or deal with it inefficiently. In our paper, we present the HairFast model, which uniquely solves these problems and achieves high resolution, near real-time performance, and superior reconstruction compared to optimization problem-based methods. Our solution includes a new architecture operating in the FS latent space of StyleGAN, an enhanced inpainting approach, and improved encoders for better alignment, color transfer, and a new encoder for post-processing. The effectiveness of our approach is demonstrated on realism metrics after random hairstyle transfer and reconstruction when the original hairstyle is transferred. In the most difficult scenario of transferring both shape and color of a hairstyle from different images, our method performs in less than a second on the Nvidia V100.
\end{abstract}    
\section{Introduction}

Advances in the generation of face images using GANs~\cite{goodfellow2014generative, karras2017progressive, karras2019style, karras2020analyzing, isola2017image, wang2018high, lee2020maskgan, jo2019sc, park2019semantic, tan2021diverse, tan2021efficient} have made it possible to apply them to semantic face editing~\cite{jiang2020psgan, portenier2018faceshop, xiao2021sketchhairsalon, yang2020deep}. One of the most challenging and interesting topic in this area is hairstyle transfer~\cite{tan2020michigan}. The essence of this task is to transfer hair attributes such as color, shape, and structure from the reference photo to the input image while preserving identity and background. The understanding of mutual interaction of these attributes is the key to a quality solution of the problem. This task has many applications among both professionals and amateurs during work with face editing programs, virtual reality and computer games. 

Existing approaches that solve this problem can be divided into two types: optimization-based~\cite{saha2021loho, zhu2021barbershop, kim2022style, khwanmuang2023stylegan, zhu2022hairnet, wei2023hairclipv2, chang2023hairnerf}, by obtaining image representations in some latent space of the image generator and directly optimizing it for the corresponding loss functions to transfer the hairstyle, and encoder-based~\cite{wei2022hairclip, karras2019style, tan2020michigan, guo2022gan}, where the whole process is done with a single direct pass through the neural network. The optimization-based methods have good quality but take too long, while the encoder-based methods are fast but still suffer from poor quality and low resolution. Moreover, both approaches still have a problem if the photos have a large pose difference.

We present a new HairFast method that works in high resolution, is outperforms in quality to state-of-the-art optimization methods, and is suitable for interactive applications in terms of speed, since we use only encoders in the inference process.
We propose our decomposition of the problem and solve each of the subtasks efficiently. In particular, we developed a new approach for pose adaptation, a new approach for FS space regularization, a more efficient approach for hair coloring, and developed a new module for detail recovery.
Our framework consists of four modules: pose alignment, shape alignment, color alignment and refinement alignment. Each module solves its own subtask by training specialized encoders. 

We have conducted an extensive series of experiments, including attribute changes both individually (color, shape) and in combination (color and shape), on the CelebA-HQ dataset~\cite{karras2017progressive} in various scenarios. Based on standard realism metrics such as FID~\cite{heusel2017gans}, $\text{FID}_{\text{CLIP}}$~\cite{Kynkaanniemi2022} and runtime, the proposed method shows comparable or even better results than state-of-the-art optimization-based methods while having inference time comparable to the fastest HairCLIP~\cite{wei2022hairclip} method.

\section{Related Works}

\paragraph{GANs.}
Generative Adversarial Networks (GANs) have significantly advanced research in image generation, and recent models such as ProgressiveGAN~\cite{karras2017progressive}, StyleGAN~\cite{karras2019style}, and StyleGAN2~\cite{karras2020analyzing} produce highly detailed and realistic images, especially in the area of human faces. Despite the progress made in face generation, high-quality, fully controlled hair editing remains a challenge due to the many side effects.

\paragraph{Latent Space Embedding.}
Inversion techniques~\cite{abdal2019image2stylegan,tewari2020pie,zhu2020domain,zhu2020improved,richardson2021encoding, tov2021designing} for StyleGAN generate latent representations that balance editability and reconstruction fidelity. Methods that prioritize editability map real images into a more flexible latent subspace, such as $W$ or $W+$~\cite{abdal2019image2stylegan}, which can reduce the accuracy of the reconstruction - a popular example is E4E~\cite{tov2021designing}. Reconstruction-focused methods, on the other hand, aim for an exact restoration of the original image. For example Barbershop merges the structural feature space ($F$) with the global style space ($S$) to form a composite space ($FS$). Such decomposition enhances the representational capacity of the space. Utilizing both the $W+$ and $FS$ latent spaces, we have created a comprehensive hair editing framework that allows for a wide range of potential realistic adjustments.

\paragraph{Optimization-based methods.}
Among the classical optimization methods, we can highlight Barbershop~\cite{zhu2021barbershop}, which uses multi-stage optimization in the StyleGAN FS space. But Barbershop doesn't work well with large pose differences, a problem that StyleYourHair~\cite{kim2022style} tries to solve by using local style matching and pose alignment loss, which allows efficient face rotation before hair transfer.
Other approaches to hair editing include: StyleGANSalon~\cite{khwanmuang2023stylegan}, which solves the rotation problem with EG3D~\cite{chan2022efficient}, HairNet~\cite{zhu2022hairnet}, which has learned to handle complex poses and uses PTI~\cite{roich2022pivotal} to improve quality but has lost the ability to independently transfer hair color, HairCLIPv2~\cite{wei2023hairclipv2} which can interact with images, masks, sketches and texts, HairNeRF~\cite{chang2023hairnerf} which uses StyleNeRF~\cite{gu2021stylenerf} instead of StyleGAN to provide distortion-free hair transfer in case of complex poses.

\paragraph{Encoder based methods.}
Encoder-based methods replace optimization processes with training a neural network, speeding up runtime a lot. Among the best models, we can highlight CtrlHair~\cite{guo2022gan} that uses SEAN~\cite{zhu2020sean} as a feature encoder and generator. The method still suffers from complex cases with different facial poses and the authors solve this by inefficient postprocessing of the mask due to which the method is slow.
HairCLIP~\cite{wei2022hairclip}, which is an order of magnitude faster than CtrlHair, uses CLIP~\cite{radford2021learning} feature extractor. The method allows to edit hair with text, but it works in $W+$ space, which causes poor preservation of face identity and hair texture.
Other approaches to hair editing include:  MichiGAN~\cite{tan2020michigan}, HairFIT~\cite{chung2022hairfit}
Encoder-based methods show significantly worse performance than optimization-based approaches especially in difficult cases with different head poses and lightning conditions. We propose the first encoder-based framework that achieves comparable quality with methods that use optimizations. 
\section{Method}

\subsection{Overview}

Formally we will solve the following problem, we have a source image $I_\text{source}$ to which we want to transfer the style and shape of $I_\text{shape}$, and an image with the desired hair color $I_\text{color}$. 

\begin{figure*}[t!]
  \centering
   \vspace{-2cm}
   \includegraphics[width=1\linewidth]{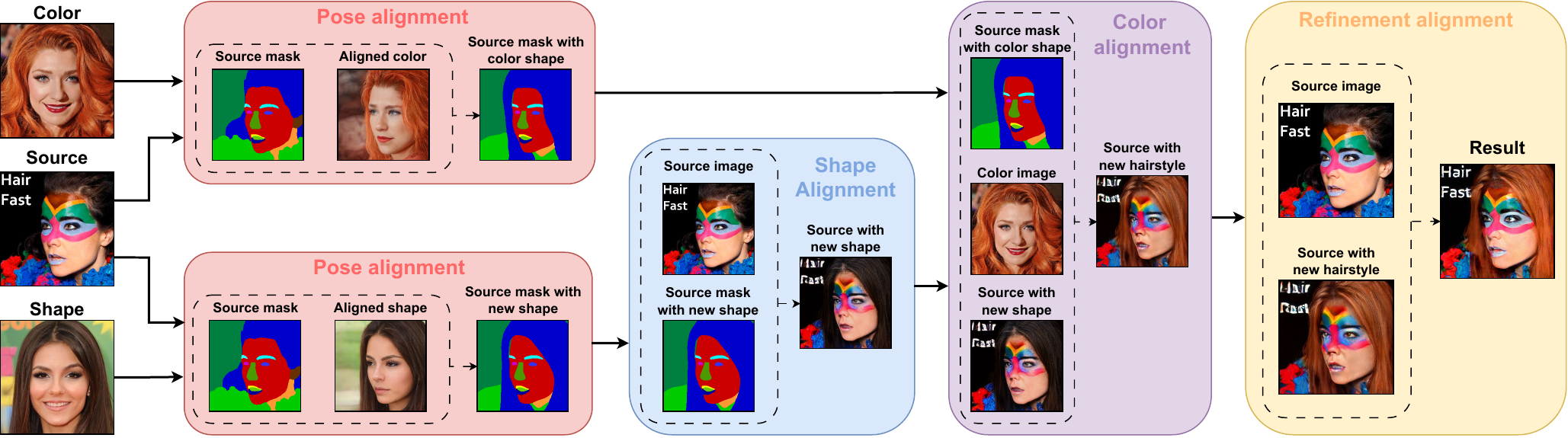}
   \caption{{\bf Overview of HairFast:} the images first pass through the Pose alignment module, which generates a pose-aligned face mask with the desired hair shape. Then we transfer the desired hairstyle shape using Shape alignment and the desired hair color using Color alignment. In the last step, Refinement alignment returns the lost details of the original image where they are needed.}
   \vspace{-0.4cm}
   \label{fig:Diagram}
\end{figure*}

In this problem setting, we will solve the shape and color transfer problems independently using Shape alignment and Color alignment respectively to make the method flexible. Also, before these modules, we propose a new stage Pose alignment which purpose is to remove the pose mismatch between image $I_\text{source}$ and images $I_\text{shape}$ and $I_\text{color}$. And after all these modules we introduce a new Refinement alignment stage, which should restore the details of the face and background that we may have lost after the hair and color transfer stage. You can see the general pipeline of our approach on the picture \cref{fig:Diagram}.

In this work, we propose a unique solution for each of these modules that shows high quality and high performance compared to existing approaches. In particular, we do not use optimization to solve these problems, which allows us to speed up the overall pipeline greatly. In our HairFast method we propose an efficient Pose alignment approach by adding a new Rotate Encoder whose purpose is to change the face latent to rotate it. In Shape Alignment we propose a new FS mixing approach that allows to transfer the hair while keeping the possibility to edit it for color changes, besides we propose an efficient use of SEAN for inpainting at this stage. Also in Color Alignment we offer a new architecture using CLIP embeddings and new losses, which has significantly improved the results of this stage. Finally, we propose a completely new Refinement alignment stage.

First of all, our method starts with Pose alignment block, its purpose is to generate a segmentation mask with a target hair shape. This block takes as input images of the original face, the desired hair and their $W+$ representations in StyleGAN space. Then, a Rotate Encoder is run inside the block, which rotates the image with the desired hair to the same pose as the original face, followed by a Shape Encoder to adapt the resulting hairstyle at the level of the segmentation mask.

In the next Shape alignment module, we transfer the hairstyle shape from $I_\text{shape}$ to $I_\text{source}$ by changing only the tensor $F$ from the FS space of StyleGAN. To do this, we generate the tensor $F$ for the inpaint after changing the shape using SEAN and target mask from Pose alignment. Once we have all the necessary $F$ tensors, we aggregate them taking into account the segmentation masks, selecting the desired parts, thus obtaining a new $F$ tensor that corresponds to the image with the desired hair shape.

The next Color module is designed to transfer the hair color from the $I_\text{color}$. To do this, we edit the $S$ space of the source image using our trained encoder, which also takes as input the $S$ tensor of the reference and additional CLIP~\cite{radford2021learning} embeddings of the source images.

The image generated after the Color module can already be considered as the final image, but in our work we also introduce a new Refinement alignment module. The purpose of this module is to restore the necessary details of the original image that were lost after inversion and editing. This allows us to preserve the identity of the face and increase the realism of the method.

\subsection{Pose alignment}

\begin{figure}[t!]
  \centering
  \vspace{-2cm}
   \includegraphics[width=1.05\linewidth]{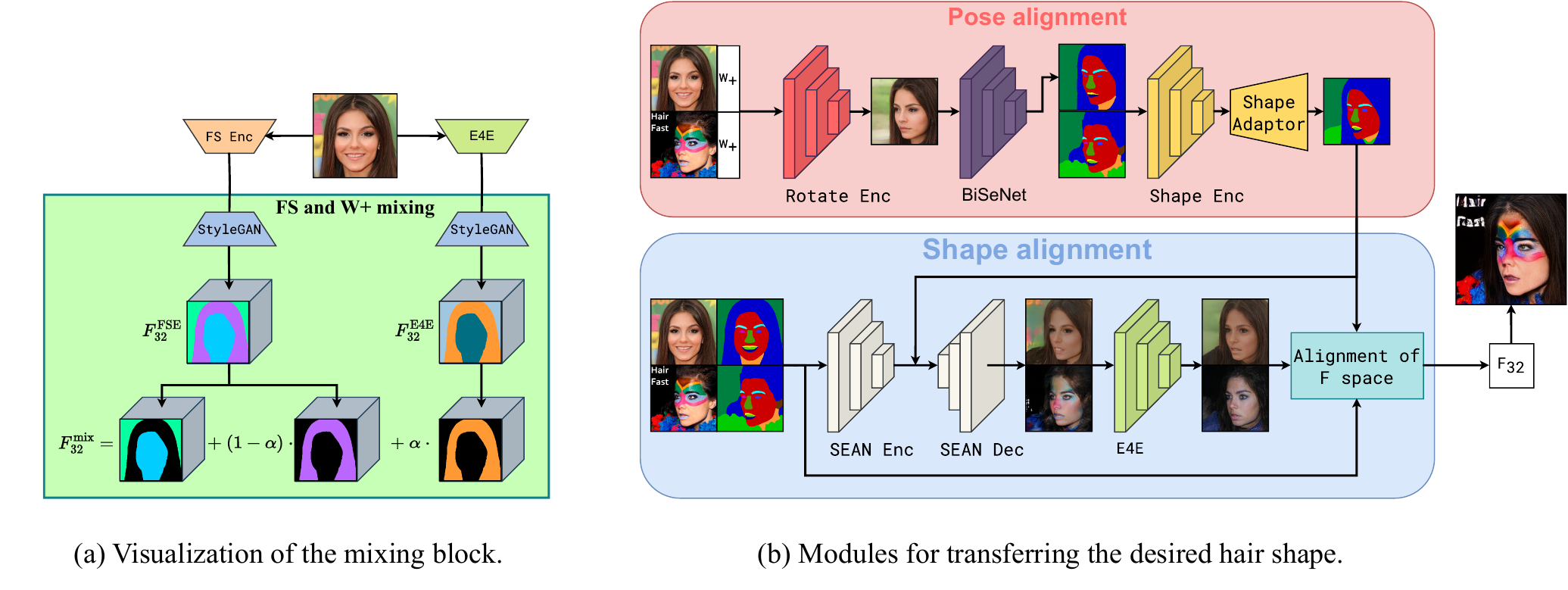}

   \caption{Detailed diagram of the units. (a) Mixing block mixes FS and W+ space representations to allow color editing (b) The Pose alignment module diagram generates a pose-aligned mask with the desired hair shape, and the Shape alignment module diagram that takes the images themselves, their segmentation masks, $W+$ and $F$ representations to transfer the desired hairstyle shape.}
   \vspace{-0.4cm}
   \label{fig:FSmixing}
   \label{fig:Alignment}
\end{figure}

In this step, we want to get a segmentation mask of the original face with the desired hair shape. Such a task of generating a target mask without optimization problems was very successfully solved by the authors of the CtrlHair~\cite{guo2022gan} method using Shape Encoder, which encodes the segmentation masks of two images as separate embeddings of hair and face, and Shape Adaptor reconstructs the segmentation mask of the desired face with the desired hair shape, additionally performing inpaint. Nevertheless, this approach still has problems.

First of all, the Shape Adaptor and Shape Encoder itself has been trained to transfer the hair shape as it is in the current pose and so for the case where the source and shape photos have too different poses, the method performs very poorly, causing the final photo to show severe hair shifts. The authors of CtrlHair have partially solved this problem with a slow and ineffective post-processing of the mask.

In our approach, we introduce a new Rotate Encoder that is trained to rotate the shape image to the same pose as the source image. This is accomplished by changing the latency of the image $w^{\text{E4E}}$ received from the E4E encoder~\cite{tov2021designing}. The new image is then segmented and given to the input of the encoder and shape adapter as the desired hair shape. Since we don't need detailed hair to generate the mask, we use the E4E representation of the image. The Rotate Encoder has been trained with very good interpolation and can rotate the image to the most complex pose while maintaining the original shape of the hair. At the same time, Encoder does not mess up the hairstyles if the image poses already match. After rotation, we obtain the image mask using BiSeNet~\cite{Yu-ECCV-BiSeNet-2018}:
\vspace{-0.1cm}
\begin{align}
 &w_{\text{rotate}} = \text{Rotate}_{\text{Enc}}(w_{\text{source}}^{\text{E4E}},\ w_{\text{shape}}^{\text{E4E}}),
 &M_\text{rotate} = \text{BiSeNet}(G(w_{\text{rotate}})),
\end{align}
\vspace{-0.6cm}

where G is StyleGAN and $w^{\text{E4E}} = \text{E4E}(I)$ -- the latent E4E representation for $I_\text{source}$ and $I_\text{shape}$.

For training Rotate Encoder, we used keypoint optimization with a pre-trained STAR~\cite{Zhou_2023_CVPR} model as well as cycle-consistency reconstruction loss. See the Appendix \ref{sec:rotate_encoder} for more information about the Rotate Encoder.

This approach allows for a high quality transfer of most hairstyles even with the most complex pose differences, correcting artifacts that occur even in the StyleYourHair method, as will be shown in the experiments section. A diagram of the Pose alignment module architecture is shown in \cref{fig:Alignment}. More formally, the generation of the final $M_\text{align} = \text{Shape}_{\text{Adaptor}}(\text{hair}_{\text{emb}},\ \text{face}_{\text{emb}})$, where

\vspace{-0.6cm}
\begin{align}
 &\text{hair}_{\text{emb}} = \text{Shape}^\text{hair}_{\text{Enc}}(M_\text{rotate}),
 &\text{face}_{\text{emb}} = \text{Shape}^\text{face}_{\text{Enc}}(M_\text{source}),\quad
  &M_\text{source} = \text{BiSeNet}(I_{\text{source}}).
\end{align}
\vspace{-0.6cm}

The last step we extract the hair area mask $H_\text{align}$ from $M_\text{align}$.

These masks $H_\text{align}$ and $M_\text{align}$ from the Pose alignment module will be needed for the next alignment tasks -- inpaint generation and shape transfer in the Shape alignment module. A diagram of the Pose alignment module is shown in \cref{fig:Alignment}.

\subsection{FS and W+ mixing}

Before transferring the new hairstyle and doing coloring, there's one more problem to address. The native integration of the FS encoder~\cite{xuyao2022} cannot edit $S$ space in a way that transfers hair color. 
To solve this problem, we additionally use E4E -- it is a very simple encoder with relatively poor image reconstruction quality, but has high editability. For this, we also reconstruct all images with E4E and mix the $F$ tensor corresponding to the hair with the $F$ tensor obtained with the FS encoder.

Formally, if $I$ is input image, then $F_{16}^{\text{FSE}},\ S = \text{FS}_{\text{Enc}}(I),\ w^{\text{E4E}} = \text{E4E}(I)$,  where $F_{16}^{\text{FSE}} \in \mathbb{R}^{16\times 16\times 512}, S \in \mathbb{R}^{12\times 512}$ -- the 
$FS$ representation obtained from FS encoder and $w^{\text{E4E}} \in \mathbb{R}^{18\times 512}$ -- the encoding from E4E.

Since we want to edit images in $F$ space of 32x32 resolution while FS encoder produces only 16x16, we need to run a few more StyleGAN blocks, while for E4E we run all 6 first blocks:

\vspace{-0.6cm}
\begin{gather}
    F_{32}^{\text{FSE}} = G_{4:6}(F_{16}^{\text{FSE}}, S), \quad\quad F_{32}^{\text{E4E}} = G_6(w^{\text{E4E}}),
\end{gather}
\vspace{-0.6cm}

where $G_6$ -- the output of the first 6 StyleGAN blocks and $G_{4:6}$ -- the generator starts at block 4.

To find the hair region in the $F$ tensor, we use BiSeNet to segment a face and downsize the selected hair mask. Then final reconstruction for images $F_{32}^{\text{mix}}$:

\vspace{-0.6cm}
\begin{align}
    M = \text{BiSeNet}(I),\quad H = \text{Downsample}_{32}(M = \text{hair}), \label{eq:hair} \\
    F_{32}^{\text{mix}} = \overline{H} \cdot F_{32}^{\text{FSE}} + (1-\alpha)\cdot H\cdot F_{32}^{\text{FSE}} + \alpha \cdot H \cdot F_{32}^{\text{E4E}}.
\end{align}
\vspace{-0.6cm}

Here $\overline{H}$ is an inversion of the mask $H$ and $\alpha$ is the hyperparameter for mixing. In our work it is equal to 0.95. This means that we only take 5\% of the hair from FS encoder, but according to our experiments, even this small mixing with the hair $F$ tensor of FS encoder hair greatly increases the quality. The visualization of the mixing procedure shown in \cref{fig:FSmixing}.

This $F_{32}^{\text{mix}}$ tensor allows us to get an excellent quality of face and background reconstruction and still edit the hairstyle.

\subsection{Shape alignment}

In this step, our goal is to transfer the desired hair shape from $I_\text{shape}$ to $I_\text{source}$. For this purpose, we edit only the $F$ space. To achieve this, we solve 2 subtasks: generation of a target mask with the desired hair shape and generation of $F$ tensor with the inpainted parts of the image.

For the inpaint task, we use the pre-trained SEAN model, which produces style vectors for each segmentation class using the input image and its segmentation mask, and its decoder reconstructs the image using the style vectors and any new segmentation mask. Thus, using this model, we can obtain a 256x256 resolution image with the desired hair shape for both source and shape photos. 

\vspace{-0.6cm}
\begin{align}
 &\text{style}_\text{codes} = \text{SEAN}_\text{Enc}(I,\ M),
 &I^\text{inpaint} = \text{SEAN}_\text{Dec}(\text{style}_\text{codes},\ M_{\text{align}}).
\end{align}
\vspace{-0.6cm}

To get the $F$ tensor representation of these images we use E4E. According to our experiments, the SEAN model in some cases produces strong artifacts in weakly represented segmentation classes such as ears, and due to artifacts on the target segmentation mask, it can also produce images with similar artifacts, such as when the hair is not connected to the head. E4E due to its good generalization is a good regularizer that automatically handles all such kind of artifact.

\vspace{-0.6cm}
\begin{align}
 &F_{32}^{\text{inpaint}} = G_6(\text{E4E}(I^\text{inpaint})).
\end{align}
\vspace{-0.6cm}

In the current step we have two initial $F_{32}^{\text{mix}}$ tensors of images and two $F_{32}^{\text{inpaint}}$ tensors after the inpaint part. The last step Alignment of F space combines all four F tensors into one new $F_{32}^\text{align}$, which can be used to generate an image with a given hairstyle. To do this, we assemble the tensor by selecting its corresponding parts using segmentation masks. A diagram of the Shape alignment module is shown in \cref{fig:Alignment}.

\vspace{-0.2cm}
\begin{equation}
    \begin{aligned}
    F_{32}^\text{align} &= H_{\text{align}} \cdot H_{\text{shape}} \cdot F^{\text{mix}}_{\text{shape}} +
     H_{\text{align}}\cdot \overline{H_{\text{shape}}} \cdot F_{\text{shape}}^{\text{inpaint}} +\\
    &+ \overline{H_{\text{align}}} \cdot \overline{H_{\text{source}}} \cdot F^{\text{mix}}_{\text{source}} +
     \overline{H_{\text{align}}} \cdot H_{\text{source}} \cdot F_{\text{source}}^{\text{inpaint}}.
    \end{aligned}
\end{equation}
\vspace{-0.25cm}

Here, $H_\text{shape}$ and $H_\text{source}$ are obtained from $I_\text{shape}$ and $I_\text{source}$ from according to \cref{eq:hair}.

\subsection{Color alignment}

In the next step, we solve the problem of changing the $S$ space so as to change the hair color to the desired color. According to our experiments, we find that learning the convex combination of $S_\text{source}$ and $S_\text{color}$ as previous methods did is not enough for good quality, so we develop a new encoder to predict the change of $S_\text{source}$. Moreover, we find that using only $S$ space is not enough to change hair color and preserve the rest of the image, so we include CLIP image embeddings to bring more information into the features. Finally, we experiment with losses and find that the cosine distance between CLIP embeddings works better than LPIPS~\cite{xuyao2022}.

As Color Encoder architecture we use 1D modulation layers similar to those used in StyleGAN. Such layers are excellent for style changing and have good stability. The purpose of this encoder is to predict the change of $S_\text{source}$ in such a way as to change the hair color to the desired color while preserving the rest of the image. We input $S_\text{color}$ as a style to the modulation layers with CLIP embeddings of the source image without hair: $\text{emb}_{\text{face}} = \text{CLIP}_{\text{enc}}(I_{\text{source}}\cdot \overline{H_{\text{align}}} \cdot \overline{H_{\text{source}}})$ and CLIP embeddings of the hair only from the color image: $\text{emb}_{\text{hair}} = \text{CLIP}_{\text{enc}}(I_\text{color}\cdot H_{\text{color}})$, where $H_{\text{align}}$ hair mask obtained from a Pose alignment module run to transfer a hair shape from $I_\text{color}$ to $I_\text{source}$. This helps to convey additional information about the original images that might have been lost after inverting images into latent spaces. And $I_{\text{blend}}$ is the final image before Refinement alignment stage:

\vspace{-0.6cm}
\begin{align}
    &S_\text{blend} = \text{Blend}_{\text{Enc}}(S_\text{source},\ S_\text{color},\ \text{emb}_{\text{face}},\ \text{emb}_{\text{hair}}),\\
    &I_{\text{blend}} = \text{StyleGAN}(F_{32}^{\text{align}},\ S_{\text{blend}}),
\end{align}
\vspace{-0.6cm}

A diagram of the method is shown in \cref{fig:Blending}.

\vspace{-0.6cm}
\begin{align}
    &\mathcal{L}_{\text{clip}}(I_\text{1}, I_\text{2}, M_\text{1}, M_\text{2}) =
    1 - \mathrm{CosSim}_\mathrm{CLIP}(I_\text{1} \cdot M_\text{1}, I_\text{2} \cdot M_\text{2}),
\end{align}
\vspace{-0.6cm}

To train the model, we use the $\mathcal{L}_{\text{clip}}$ one of which optimizes the cosine distance between the CLIP embeddings of the final and source images on both the reconstruction and transfer of the desired color. See the Appendix \ref{sec:blending_encoder} for training and encoder details.

\subsection{Refinement alignment}

\begin{figure}[t!]
  \centering
   \vspace{-2cm}
   \includegraphics[width=1\linewidth]{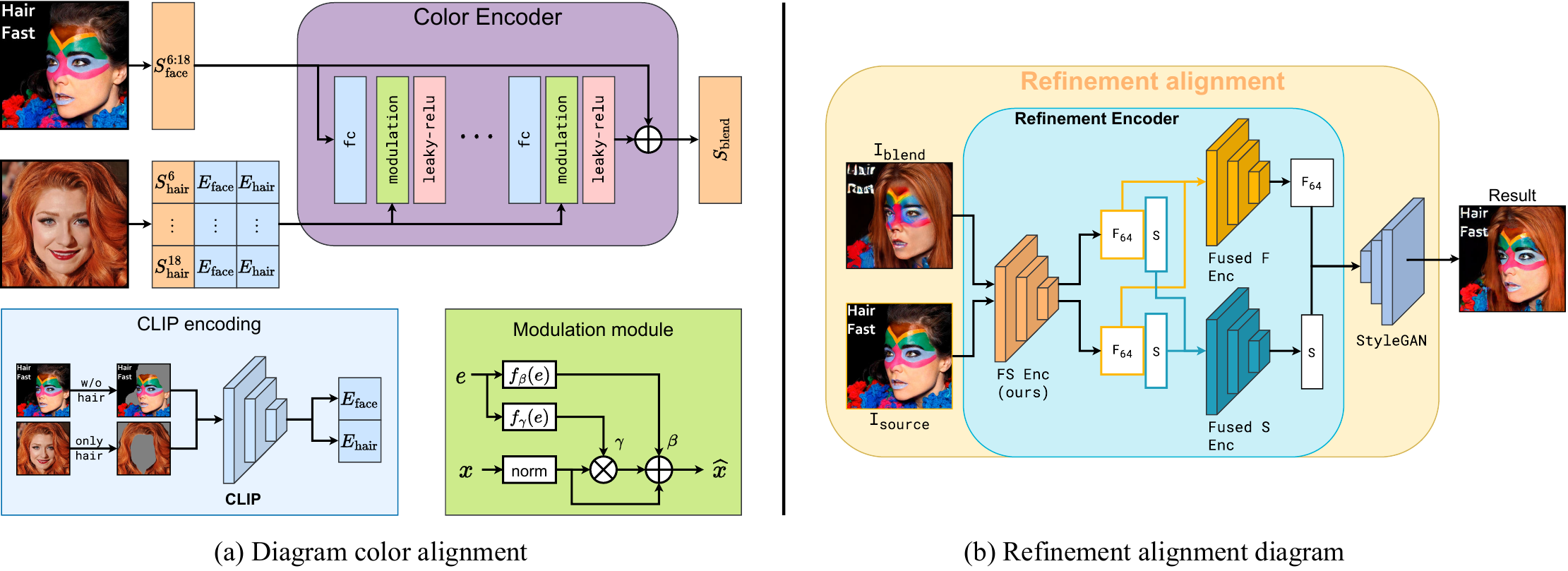}
   \caption{Detailed diagram of the units. (a) A color alignment module diagram that takes as input $S$ image representations as well as segmentation masks. The purpose of this block is to encode the details of the original image and change the $S$ space to transfer the desired hair color and preserve the identity. (b) A refinement alignment diagram that takes as input the source image and post Color alignment module image. At this module, the goal is to get a new representation in StyleGAN space to get a realistic image, with the original details of the source image that were lost after inverting images into latents.}
   \vspace{-0.3cm}
   \label{fig:Blending}
   \label{fig:PostProcess}
\end{figure}

Our method, even though it has a higher quality Color alignment step, still has a problem on complex cases where the face hue may change. Particularly because of this we cannot simply use Poisson blending like a CtrlHair, as the difference in shades emphasizes the overlay more and visually it doesn't look realistic.

For this reason, we are developing our own Refinement alignment module, which is essentially a larger and more powerful reconstruction encoder, but for a more complex task -- reconstruction of the original face and background, reconstruction of hair after Color alignment module and inpaint of non-matching parts. This Refinement encoder generates an F tensor 4 times higher resolution than the FS encoder we used in the Shape alignment module. This allows for unrivaled reconstruction quality. Unlike traditional encoders that sacrifice reconstruction quality for good editing, we are able to use such a large resolution F tensor due to the fact that we do not have to edit the image after this module.

Refinement itself consists of a trained FS encoder at resolution 64 for the usual reconstruction task, we use it to encode the original image $I_{\text{source}}$ and the $I_{\text{blend}}$ image after our method:

\vspace{-0.6cm}
\begin{align}
    F_{64}^{\text{blend}},\ S^{\text{blend}} &= FS_{\text{Enc (ours)}}(I_{\text{blend}}),\\
    F_{64}^{\text{source}},\ S^{\text{source}} &= FS_{\text{Enc (ours)}}(I_{\text{source}}).
\end{align}
\vspace{-0.6cm}

The resulting tensors F are fused using IResNet blocks. In turn, S space is fused using two similar Color Encoder models, but without additional CLIP features. The output of this composite encoder is $F_{64}^{\text{final}} = \text{Fused}_{\text{F Enc}}(F_{64}^{\text{blend}},\ F_{64}^{\text{source}})$ tensor and an $S_{\text{final}} = \text{latent}_{\text{avg}} + \text{Fused}_{\text{S Enc}}(S^{\text{blend}},\ S^{\text{source}})$ vector, which are input to StyleGAN to generate the final image $I_{\text{final}}$:

\vspace{-0.6cm}
\begin{align}
    &I_{\text{final}} = \text{StyleGAN}(F_{64}^{\text{final}},\ S_{\text{final}}).
\end{align}
\vspace{-0.6cm}

For model training, we use loss functions for hair reconstruction and original parts of the image, and for inpaint we use guidelines from the more robust StyleGAN space and adversarial loss. For reconstruction, these include multi-scale perceptual loss~\cite{xuyao2022}, DSC++~\cite{yeung2023calibrating}, ArcFace and regularizations. See the Appendix \ref{sec:post_processing} for training and encoders details.

A diagram of the Refinement procedure is shown in the \cref{fig:PostProcess}. This produces the final HairFast model, which is shown in the \cref{fig:Diagram}.

\section{Experiments}\label{sec:experimets}

\textbf{Realism after editing.} The task of hairstyle transfer is very challenging, largely due to the lack of objective metrics. One possible metric to reflect the quality of hairstyle transfer is to measure the realism of the image using FID. To measure this metric, we consider 4 main cases: transferring hairstyle and color from different images (full), transferring only a new hairstyle shape (shape), transferring only a new color (color), and transferring both color and shape from the same image (both). To measure the metrics, we use the CelebA-HQ~\cite{karras2017progressive} dataset, from which we capture 1000 to 3000 experiments for each case, on which we run all methods.
We used methods such as HairCLIP~\cite{wei2022hairclip}, HairCLIPv2~\cite{wei2023hairclipv2}, CtrlHair~\cite{guo2022gan}, StyleYourHair~\cite{kim2022style} and Barbershop~\cite{zhu2021barbershop} for comparison, and for their inference we used the official code implementation.
Additionally, we measure the median running time among all runs of these experiments, excluding the time to save the results to disk and initialize the neural networks. In the \cref{table:table_FID_final} you can observe the results of this experiment for the “full” and “both” cases, the complete table can be seen in the Appendix \ref{sec:full_metrics}.

In these experiments, we do not compare with HairNet~\cite{zhu2022hairnet}, HairNeRF~\cite{chang2023hairnerf} and StyleGANSalon~\cite{khwanmuang2023stylegan} due to the lack of their code and the inability to run the methods on our images. Instead, we compare with StyleGANSalon on images they published from their inference along with LOHO~\cite{saha2021loho} in Appendix \ref{sec:salon_compare}, and we also make a visual comparison with HairNet and HairNeRF on images from their article in Appendix \ref{sec:visual_compare_hairnet}.

\begin{table*}[t!]
\centering
\vspace{-2cm}
\caption{\textbf{Realism Metrics}. These metrics were measured on the same pre-selected triples of images (face, shape and color) from the CelebaHQ~\cite{karras2017progressive} dataset. Then, applying the method, FID was measured on the original dataset and the modified dataset. $\text{FID}_{\text{CLIP}}$~\cite{Kynkaanniemi2022} was counted similarly to FID, but a CLIP encoder was used instead of Inception V3. Running time was measured as the median time among a bunch of method runs, without taking into account loading images from disk. \textbf{Pose Metrics}. For this metrics, we consider color and shape transfer from the target image to the source image. We divided all pairs into 3 equal buckets: easy, medium and hard according to the difference of face key points. \textbf{Reconstruction}. For this each method is started on the task of transferring the color and shape of the hairstyle from itself to itself, thus at the end we measure the metrics with the original image.}
\resizebox{\textwidth}{!}{

\begin{tabular}{l*{12}{c}}
\toprule
\multirow{3}{*}{\textbf{Model}} & \multicolumn{4}{c}{\textbf{Realism metrics}} & \multicolumn{4}{c}{\textbf{Pose metrics}} & \multicolumn{2}{c}{\textbf{Reconstruction}} \\
\cmidrule(lr){2-5}
\cmidrule(lr){6-9}
\cmidrule(lr){10-11}
& \multicolumn{2}{c}{FID↓} & \multicolumn{2}{c}{$\text{FID}_{\text{CLIP}}$↓} & \multicolumn{2}{c}{FID↓} & \multicolumn{2}{c}{$\text{FID}_{\text{CLIP}}$↓} & LPIPS↓ & PSNR↑ & \multicolumn{2}{c}{Time (s)↓} \\
\cmidrule(lr){2-3}
\cmidrule(lr){4-5}
\cmidrule(lr){6-7}
\cmidrule(lr){8-9}
\cmidrule(lr){10-11}
\cmidrule(lr){12-13}

                                    & full                & both                & full               & both               & medium              & hard                & medium              & hard                &               &                & A100               & V100 \\              
\midrule
HairCLIP~\cite{wei2022hairclip}     & 34.95               & 40.68               & 12.20              & 13.32              & 55.77               & 54.35               & 15.53               & 15.73               & 0.36               & 14.08               & \boldmath$0.28$    & \boldmath$0.36$ \\   
HairCLIPv2~\cite{wei2023hairclipv2} & $\underline{14.28}$ & $\underline{23.37}$ & 10.98              & 12.14              & 44.62               & $\underline{45.28}$ & 14.56               & 18.66               & 0.16               & 19.71               & 112                & 221 \\               
CtrlHair~\cite{guo2022gan}          & $15.10$             & 24.81               & 9.52               & 10.42              & 46.45               & 50.12               & 12.96               & 16.42               & 0.15               & 19.96               & 6.57               & 7.87 \\              
StyleYourHair~\cite{kim2022style}   & -                   & 25.90               & -                  & 10.91              & 46.32               & 47.19               & 13.70               & 15.93               & 0.14               & $\underline{21.74}$ & 84                 & 239 \\               
Barbershop~\cite{zhu2021barbershop} & 15.94               & 24.52               & $\underline{7.07}$ & $\underline{8.12}$ & $\underline{44.08}$ & 46.13               & $\underline{11.27}$ & $\underline{13.30}$ & $\underline{0.11}$ & 21.18               & 213                & 645 \\               
HairFast (ours)                     & \boldmath$13.12$    & \boldmath$22.71$    & \boldmath$5.12$    & \boldmath$6.06$    & \boldmath$43.25$    & \boldmath$44.85$    & \boldmath$8.90$     & \boldmath$10.33$    & \boldmath$0.08$    & \boldmath$23.45$    & $\underline{0.41}$ & $\underline{0.78}$ \\
\bottomrule
\end{tabular}

}
\label{table:table_FID_final}
\label{table:pose_diff}
\label{table:reconstruct}
\vspace{-0.1cm}
\end{table*}

\begin{table*}[t!]
\centering
\vspace{-2cm}
\caption{A comparison of the characteristics of the main hair transfer methods.}
    \resizebox{\textwidth}{!}{
        \begin{tabular}{l*{9}{@{\hskip 0.25cm}c}}
        \toprule
                                   & \makecell[c]{Barbershop\\~\cite{zhu2021barbershop}} & \makecell[c]{StyleYourHair\\~\cite{kim2022style}} & \makecell[c]{HairNet\\~\cite{zhu2022hairnet}} & \makecell[c]{HairNeRF\\~\cite{chang2023hairnerf}} & \makecell[c]{StyleSalon\\~\cite{khwanmuang2023stylegan}} & \makecell[c]{CtrlHair\\~\cite{guo2022gan}} & \makecell[c]{HairCLIP\\~\cite{wei2022hairclip}} & \makecell[c]{HairCLIPv2\\~\cite{wei2023hairclipv2}} & Ours\\

        \midrule
        \textbf{Quality}\\
            Hair realism                  & \color{Green}{High}                 & \color{Green}{High}               & \color{Green}{High}           & \color{Green}{High}               & \color{Green}{High}                      & \color{YellowOrange}{Medium} & \color{Red}{Low}                & \color{YellowOrange}{Medium}        & \color{Green}{High}    \\[0.1cm]
            \makecell[l]{Face-background\\preservation}  & \color{YellowOrange}{Medium}          & \color{YellowOrange}{Medium}        & \color{YellowOrange}{Medium}    & \color{Green}{High}                & \color{Green}{High}                          & \color{Green}{High}          & \color{Red}{Low}                 & \color{YellowOrange}{Medium}          & \color{Green}{High}    \\
        
            \midrule 
            \textbf{Functionality}\\                                                                                                                                                                                                                                                                                                                  
            Pose alignment                & \color{Red}\ding{55}                & \color{Green}\ding{51}            & \color{Green}\ding{51}        & \color{Green}\ding{51}            & \color{Green}\ding{51}                      & \color{Red}\ding{55}       & \color{Red}\ding{55}            & \color{Green}\ding{51}              & \color{Green}\ding{51} \\[0.1cm]
            \makecell[l]{Separate shape/\\color transfer} & \color{Green}\ding{51}              & \color{Red}\ding{55}              & \color{Red}\ding{55}          & \color{Red}\ding{55}              & \color{Red}\ding{55}                        & \color{Green}\ding{51}     & \color{Green}\ding{51}          & \color{Green}\ding{51}              & \color{Green}\ding{51} \\

            \midrule
            \textbf{Efficiency}\\
            W/o optimization          & \color{Red}\ding{55}                & \color{Red}\ding{55}              & \color{Red}\ding{55}          & \color{Red}\ding{55}              & \color{Red}\ding{55}                        & \color{Green}\ding{51}     & \color{Green}\ding{51}          & \color{Red}\ding{55}                & \color{Green}\ding{51} \\[0.1cm]
            Runtime                       & \color{Red}{>10m}                   & \color{Red}{>3m}         & \color{Red}{>3m}     & \color{Red}{>3m}         & \color{Red}{>10m}                           & \color{YellowOrange}{>5s}   &  \color{Green}{<1s}             & \color{Red}{>3m}           & \color{Green}{<1s}     \\
        
            \midrule
            \textbf{Reproducibility}\\
            Code access                   & \color{Green}\ding{51}              & \color{Green}\ding{51}            & \color{Red}\ding{55}          & \color{Red}\ding{55}              & \color{Red}\ding{55}                        & \color{Green}\ding{51}     & \color{Green}\ding{51}          & \color{Green}\ding{51}              & \color{Green}\ding{51} \\
        \bottomrule
        \end{tabular}
    }
    \label{table:table_characterization}
\vspace{-0cm}
\end{table*}

As we can see in the \cref{table:table_FID_final}, a method like CtrlHair outperforms optimization-based methods like Barbershop and StyleYourHair by FID metrics. However, visual analysis reveals that the method performs much worse and artifacts are visible, which appear as a consequence of strong Poisson Blending of the final image with the original image. The authors in~\cite{Kynkaanniemi2022} studied the problem that makes images with strong artifacts appear more realistic by FID metric. They were able to solve this problem by using the $\text{FID}_{\text{CLIP}}$ metric, which simply uses higher quality embeddings from the CLIP model. We also compute this metric in our experiments. Note that the metric uses the CLIP-ViT-B-32 checkpoint while we use CLIP-VIT-B-16 for color encoder training, so there is no leakage in our measurements.

Analyzing the results, our method performs better on all metrics. Looking at runtime, we outperform Barbershop on V100 by a factor of 800 and even CtrlHair by more than 10 times. This is because CtrlHair has an expensive post-processing implementation for alignment and Poisson blending. The only method that is faster is HairCLIP, but its performance in our problem setup is quite poor.

\textbf{Pose difference}. \cref{table:pose_diff} shows the results of the metrics on a subsample of our main experiment "both", but split into different cases of pose difference. For this purpose, we counted the RMSE of key points of the source image with the shape image and split all cases of hairstyle transfer into 3 equal folds: easy, medium and hard. The last two cases are presented in the table. The full table can be seen in the Appendix \ref{sec:full_metrics}.

\textbf{Reconstruction}. Another quality metric can be the reconstruction metric, where each method tries to transfer the shape and color of the hairstyle from itself, in this case we have a ground truth image with which we can measure the metrics. For this metric, we measure the LPIPS and PSNR between the original images and the resulting images. 3000 random images from CelebA-HQ were taken for reconstruction. Additional metrics for reconstruction such as $\text{FID}$ and $\text{FID}_\text{CLIP}$ can be viewed in the Appendix \ref{sec:full_metrics}.

Analyzing the results \cref{table:reconstruct}, we also outperform other methods. This confirms the effectiveness of our Refinement alignment stage, which recovers lost image details during encoding, outperforming even optimization-based methods.

\begin{figure*}[t!]
  \centering
  \vspace{-0.3cm}
   \includegraphics[width=1\linewidth]{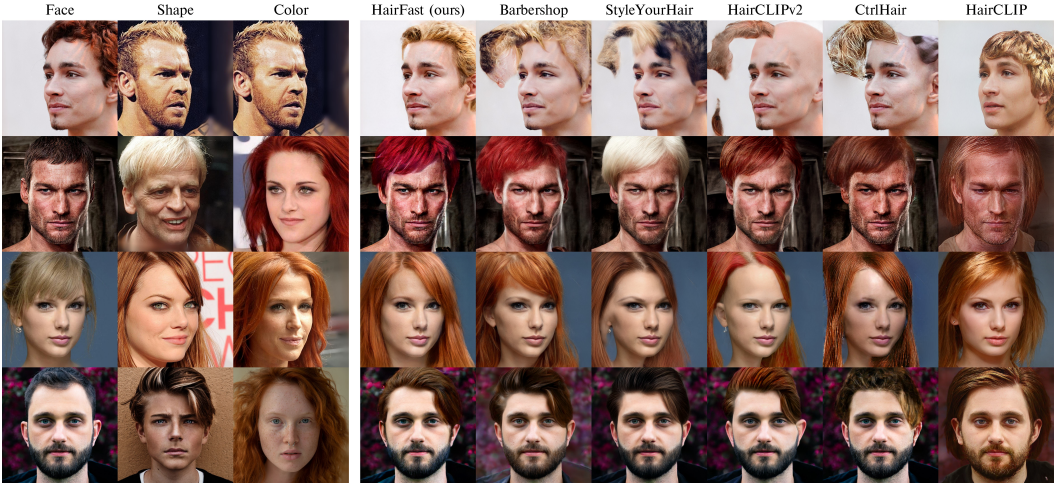}

   \caption{Visual comparison of methods on different cases for transferring hair and color together, or separately. StyleYourHair transfers color only from the Shape image. According to the results of visual comparison, our model better preserves the identity of the source image. At the same time, our method on most cases better transfers the desired hair color and texture, and works better with complex pose differences. For a more detailed comparison, see Appendix \ref{sec:visual_comp}.}
   \label{fig:MainCompare}
   \vspace{-0.1cm}
\end{figure*}

\begin{table}[t!]
\vspace{-2cm}
\centering
\def\arraystretch{1.2}
\begin{minipage}{0.35\linewidth}\centering
\resizebox{\columnwidth}{!}{%
\begin{tabular}{lccc}
        \toprule
            \textbf{\quad \ \ \ Configuration}         & FID↓ & $\text{FID}_{\text{CLIP}}$↓ & Time (s)↓ \\
        \midrule
            A\quad Baseline & 16.23 & 6.92 & 0.67 \\
        \midrule
            B\quad w/o Color alignment & 26.88 & 11.45 & 39 \\
            C\quad w/o F mix. & 16.57 & 6.72 & $\underline{0.67}$ \\
            D\quad config B w/o F mix. & 27.74 & 11.51 & 39 \\
        \midrule
            E\quad w/o Rotate Encoder & 16.87 & 7.52 & ${\bf 0.62}$ \\
            F\quad w/o Shape Adaptor & 18.72 & 6.37 & 37 \\
            G\quad w/o SEEN inpaint & ${\bf 12.79}$ & ${\bf 4.58}$ & 92 \\
        \midrule
            H\quad + Refinement (ours) & $\underline{13.12}$ & $\underline{5.12}$ & 0.78 \\
        \bottomrule
        \end{tabular}%
        }
\vspace{-0.1cm}
\caption{\small Ablation results. Baseline is HairFast, but without Refinement alignment. In each configuration, we replace the specified part with optimizations from Barbershop as needed.}
        \label{table:ablations}
\end{minipage}
\quad
\begin{minipage}{0.59\linewidth}\centering
\centering
 \vspace{-0.1cm}
      \raisebox{0.7cm}{\includegraphics[width=\linewidth]{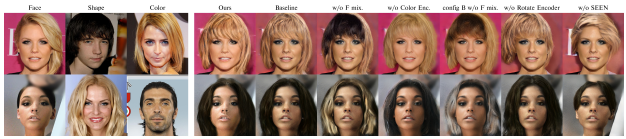}}
      \vspace{-0.32cm}\captionof{figure}{\small Ablation Study for different configurations of our model. Our model is used as the Baseline, but without Refinement alignment. Each column represents a change in the Baseline of the model.}
      \label{fig:Ablations}
      
\end{minipage}
\vspace{-0.6cm}
\end{table}

\textbf{Overall comparison.} The \cref{table:table_characterization} shows a comparison of the characteristics of the methods. Hair realism was determined according to realism metrics on reconstruction tasks and visual comparison. HairCLIPv2 has medium realism because of poor reconstruction, which due to the peculiarities of the architecture does not allow to transfer the desired texture accurately enough, in turn, CtrlHair despite the excellent metrics in visual comparison shows not similar to the desired results due to the limitations of the generator. The other methods, except HairCLIP, transfer the hairstyle realistically. 

When it comes to preserving face and background details, the latent space of methods in which image inversions take place is mainly responsible for this. Methods such as Barbershop, StyleYourHair, HairNet and HairCLIPv2 use FS resolution space 32, which does not allow them to preserve much details. In turn, the HairNeRF and StyleGANSalon methods use PTI, which allows them to preserve more details of the original image, and the CtrlHair method uses Poisson Blending, which also allows direct transfer of all original details. Our method uses FS resolution space 64, which when compared visually and reconstruction metrics shows even better quality than methods with PTI. HairCLIP, in contrast, uses the weakest W+ space. 

The runtime of each method that has a code we tested on the Nvidia V100. StyleGANSalon's time estimate came from their article, where they claim to run longer than Barbershop, while methods like HairNet and HairNeRF use PTI which makes them take at least a few minutes per image. 

Looking at the rest of the features, unlike some other methods we are able to transfer hair color and shape independently, we are also able to handle large pose differences and our entire architecture consists of encoders, which allows us to work very fast. Moreover our method has code for inference, all pre-trained weights and scripts for training for full reproducibility.

\textbf{Ablation study.} As ablation, we remove some parts of our method and replace them with Barbershop optimization processes if necessary. On these configurations we measure the realism metrics after the hairstyle transfer, the results of which can be seen in the table Tab. \ref{table:ablations} and images \cref{fig:Ablations}.

By ablation, we proved the high quality of Color Encoder, the necessity of mixing FS and W+ spaces, the effectiveness of Rotate Encoder, Shape Adaptor, SEEN, and the effectiveness of Refinement alignment module. For more detailed ablation conclusions, see Appendix \ref{sec:ablation_detail}.

\textbf{Failure cases.} The main problems of our method arise in the inpaint part, it may not work well when long hair is replaced by short hair. Also, our method suffers from transferring hair with complex textures such as ponytails, ribbons and braids. While these problems are important, they are inherent in all baseline models and we will address them in our future work. A detailed analysis of failure cases can be seen in the Appendix \ref{sec:failure_cases}.
\section{Conclusion and Limitations}

In this article, we introduced the new HairFast method for hair transfer. Unlike other approaches, we were able to achieve high quality and high resolution outperforms to other optimization-based methods, but still working in near real time. 
We developed a new approach for pose adaptation, a new approach for FS space regularization, a more efficient approach for hair coloring, and developed a new module for detail recovery.

But our method, like many others, is limited by the small number of ways to transfer hairstyles, but our architecture allows to fix this in future work. For example, our Color alignment module architecture allows similarly to HairCLIP to do hair color editing with text, and using Shape Adaptor allows similarly to CtrlHair to edit hair shape with sliders.

We prove the effectiveness of our approach by comparing it with other methods in the \cref{sec:experimets}, and refer to additional experiments in the Appendix \ref{sec:full_metrics}, \ref{sec:salon_compare}, \ref{sec:visual_compare_hairnet}, \ref{sec:color_transfer} and \ref{sec:visual_comp} for further details.

\bibliographystyle{plain}
{
    \small
    \bibliography{main}

\begin{thebibliography}{10}

\bibitem{abdal2019image2stylegan}
Rameen Abdal, Yipeng Qin, and Peter Wonka.
\newblock Image2stylegan: How to embed images into the stylegan latent space?
\newblock In {\em Proceedings of the IEEE/CVF international conference on computer vision}, pages 4432--4441, 2019.

\bibitem{chan2022efficient}
Eric~R Chan, Connor~Z Lin, Matthew~A Chan, Koki Nagano, Boxiao Pan, Shalini De~Mello, Orazio Gallo, Leonidas~J Guibas, Jonathan Tremblay, Sameh Khamis, et~al.
\newblock Efficient geometry-aware 3d generative adversarial networks.
\newblock In {\em Proceedings of the IEEE/CVF Conference on Computer Vision and Pattern Recognition}, pages 16123--16133, 2022.

\bibitem{chang2023hairnerf}
Seunggyu Chang, Gihoon Kim, and Hayeon Kim.
\newblock Hairnerf: Geometry-aware image synthesis for hairstyle transfer.
\newblock In {\em Proceedings of the IEEE/CVF International Conference on Computer Vision}, pages 2448--2458, 2023.

\bibitem{chung2022hairfit}
Chaeyeon Chung, Taewoo Kim, Hyelin Nam, Seunghwan Choi, Gyojung Gu, Sunghyun Park, and Jaegul Choo.
\newblock Hairfit: pose-invariant hairstyle transfer via flow-based hair alignment and semantic-region-aware inpainting.
\newblock {\em arXiv preprint arXiv:2206.08585}, 2022.

\bibitem{goodfellow2014generative}
Ian Goodfellow, Jean Pouget-Abadie, Mehdi Mirza, Bing Xu, David Warde-Farley, Sherjil Ozair, Aaron Courville, and Yoshua Bengio.
\newblock Generative adversarial nets.
\newblock {\em Advances in neural information processing systems}, 27, 2014.

\bibitem{gu2021stylenerf}
Jiatao Gu, Lingjie Liu, Peng Wang, and Christian Theobalt.
\newblock Stylenerf: A style-based 3d-aware generator for high-resolution image synthesis, 2021.

\bibitem{guo2022gan}
Xuyang Guo, Meina Kan, Tianle Chen, and Shiguang Shan.
\newblock Gan with multivariate disentangling for controllable hair editing.
\newblock In {\em European Conference on Computer Vision}, pages 655--670. Springer, 2022.

\bibitem{heusel2017gans}
Martin Heusel, Hubert Ramsauer, Thomas Unterthiner, Bernhard Nessler, and Sepp Hochreiter.
\newblock Gans trained by a two time-scale update rule converge to a local nash equilibrium.
\newblock {\em Advances in neural information processing systems}, 30, 2017.

\bibitem{isola2017image}
Phillip Isola, Jun-Yan Zhu, Tinghui Zhou, and Alexei~A Efros.
\newblock Image-to-image translation with conditional adversarial networks.
\newblock In {\em Proceedings of the IEEE conference on computer vision and pattern recognition}, pages 1125--1134, 2017.

\bibitem{jiang2020psgan}
Wentao Jiang, Si~Liu, Chen Gao, Jie Cao, Ran He, Jiashi Feng, and Shuicheng Yan.
\newblock Psgan: Pose and expression robust spatial-aware gan for customizable makeup transfer.
\newblock In {\em Proceedings of the IEEE/CVF Conference on Computer Vision and Pattern Recognition}, pages 5194--5202, 2020.

\bibitem{jo2019sc}
Youngjoo Jo and Jongyoul Park.
\newblock Sc-fegan: Face editing generative adversarial network with user's sketch and color.
\newblock In {\em Proceedings of the IEEE/CVF international conference on computer vision}, pages 1745--1753, 2019.

\bibitem{karras2017progressive}
Tero Karras, Timo Aila, Samuli Laine, and Jaakko Lehtinen.
\newblock Progressive growing of gans for improved quality, stability, and variation.
\newblock {\em arXiv preprint arXiv:1710.10196}, 2017.

\bibitem{karras2019style}
Tero Karras, Samuli Laine, and Timo Aila.
\newblock A style-based generator architecture for generative adversarial networks.
\newblock In {\em Proceedings of the IEEE/CVF conference on computer vision and pattern recognition}, pages 4401--4410, 2019.

\bibitem{karras2020analyzing}
Tero Karras, Samuli Laine, Miika Aittala, Janne Hellsten, Jaakko Lehtinen, and Timo Aila.
\newblock Analyzing and improving the image quality of stylegan.
\newblock In {\em Proceedings of the IEEE/CVF conference on computer vision and pattern recognition}, pages 8110--8119, 2020.

\bibitem{khwanmuang2023stylegan}
Sasikarn Khwanmuang, Pakkapon Phongthawee, Patsorn Sangkloy, and Supasorn Suwajanakorn.
\newblock Stylegan salon: Multi-view latent optimization for pose-invariant hairstyle transfer.
\newblock In {\em Proceedings of the IEEE/CVF Conference on Computer Vision and Pattern Recognition}, pages 8609--8618, 2023.

\bibitem{kim2022style}
Taewoo Kim, Chaeyeon Chung, Yoonseo Kim, Sunghyun Park, Kangyeol Kim, and Jaegul Choo.
\newblock Style your hair: Latent optimization for pose-invariant hairstyle transfer via local-style-aware hair alignment.
\newblock In {\em European Conference on Computer Vision}, pages 188--203. Springer, 2022.

\bibitem{Kynkaanniemi2022}
Tuomas Kynkäänniemi, Tero Karras, Miika Aittala, Timo Aila, and Jaakko Lehtinen.
\newblock The role of imagenet classes in fréchet inception distance.
\newblock In {\em Proc. ICLR}, 2023.

\bibitem{lee2020maskgan}
Cheng-Han Lee, Ziwei Liu, Lingyun Wu, and Ping Luo.
\newblock Maskgan: Towards diverse and interactive facial image manipulation.
\newblock In {\em Proceedings of the IEEE/CVF Conference on Computer Vision and Pattern Recognition}, pages 5549--5558, 2020.

\bibitem{park2019semantic}
Taesung Park, Ming-Yu Liu, Ting-Chun Wang, and Jun-Yan Zhu.
\newblock Semantic image synthesis with spatially-adaptive normalization.
\newblock In {\em Proceedings of the IEEE/CVF conference on computer vision and pattern recognition}, pages 2337--2346, 2019.

\bibitem{portenier2018faceshop}
Tiziano Portenier, Qiyang Hu, Attila Szabo, Siavash~Arjomand Bigdeli, Paolo Favaro, and Matthias Zwicker.
\newblock Faceshop: Deep sketch-based face image editing.
\newblock {\em arXiv preprint arXiv:1804.08972}, 2018.

\bibitem{radford2021learning}
Alec Radford, Jong~Wook Kim, Chris Hallacy, Aditya Ramesh, Gabriel Goh, Sandhini Agarwal, Girish Sastry, Amanda Askell, Pamela Mishkin, Jack Clark, et~al.
\newblock Learning transferable visual models from natural language supervision.
\newblock In {\em International conference on machine learning}, pages 8748--8763. PMLR, 2021.

\bibitem{richardson2021encoding}
Elad Richardson, Yuval Alaluf, Or~Patashnik, Yotam Nitzan, Yaniv Azar, Stav Shapiro, and Daniel Cohen-Or.
\newblock Encoding in style: a stylegan encoder for image-to-image translation.
\newblock In {\em Proceedings of the IEEE/CVF conference on computer vision and pattern recognition}, pages 2287--2296, 2021.

\bibitem{roich2022pivotal}
Daniel Roich, Ron Mokady, Amit~H Bermano, and Daniel Cohen-Or.
\newblock Pivotal tuning for latent-based editing of real images.
\newblock {\em ACM Transactions on graphics (TOG)}, 42(1):1--13, 2022.

\bibitem{saha2021loho}
Rohit Saha, Brendan Duke, Florian Shkurti, Graham~W Taylor, and Parham Aarabi.
\newblock Loho: Latent optimization of hairstyles via orthogonalization.
\newblock In {\em Proceedings of the IEEE/CVF Conference on Computer Vision and Pattern Recognition}, pages 1984--1993, 2021.

\bibitem{tan2021diverse}
Zhentao Tan, Menglei Chai, Dongdong Chen, Jing Liao, Qi~Chu, Bin Liu, Gang Hua, and Nenghai Yu.
\newblock Diverse semantic image synthesis via probability distribution modeling.
\newblock In {\em Proceedings of the IEEE/CVF Conference on Computer Vision and Pattern Recognition}, pages 7962--7971, 2021.

\bibitem{tan2020michigan}
Zhentao Tan, Menglei Chai, Dongdong Chen, Jing Liao, Qi~Chu, Lu~Yuan, Sergey Tulyakov, and Nenghai Yu.
\newblock Michigan: multi-input-conditioned hair image generation for portrait editing.
\newblock {\em arXiv preprint arXiv:2010.16417}, 2020.

\bibitem{tan2021efficient}
Zhentao Tan, Dongdong Chen, Qi~Chu, Menglei Chai, Jing Liao, Mingming He, Lu~Yuan, Gang Hua, and Nenghai Yu.
\newblock Efficient semantic image synthesis via class-adaptive normalization.
\newblock {\em IEEE Transactions on Pattern Analysis and Machine Intelligence}, 44(9):4852--4866, 2021.

\bibitem{tewari2020pie}
Ayush Tewari, Mohamed Elgharib, Florian Bernard, Hans-Peter Seidel, Patrick P{\'e}rez, Michael Zollh{\"o}fer, and Christian Theobalt.
\newblock Pie: Portrait image embedding for semantic control.
\newblock {\em ACM Transactions on Graphics (TOG)}, 39(6):1--14, 2020.

\bibitem{tov2021designing}
Omer Tov, Yuval Alaluf, Yotam Nitzan, Or~Patashnik, and Daniel Cohen-Or.
\newblock Designing an encoder for stylegan image manipulation.
\newblock {\em ACM Transactions on Graphics (TOG)}, 40(4):1--14, 2021.

\bibitem{wang2018high}
Ting-Chun Wang, Ming-Yu Liu, Jun-Yan Zhu, Andrew Tao, Jan Kautz, and Bryan Catanzaro.
\newblock High-resolution image synthesis and semantic manipulation with conditional gans.
\newblock In {\em Proceedings of the IEEE conference on computer vision and pattern recognition}, pages 8798--8807, 2018.

\bibitem{wei2022hairclip}
Tianyi Wei, Dongdong Chen, Wenbo Zhou, Jing Liao, Zhentao Tan, Lu~Yuan, Weiming Zhang, and Nenghai Yu.
\newblock Hairclip: Design your hair by text and reference image.
\newblock In {\em Proceedings of the IEEE/CVF Conference on Computer Vision and Pattern Recognition}, pages 18072--18081, 2022.

\bibitem{wei2023hairclipv2}
Tianyi Wei, Dongdong Chen, Wenbo Zhou, Jing Liao, Weiming Zhang, Gang Hua, and Nenghai Yu.
\newblock Hairclipv2: Unifying hair editing via proxy feature blending, 2023.

\bibitem{xiao2021sketchhairsalon}
Chufeng Xiao, Deng Yu, Xiaoguang Han, Youyi Zheng, and Hongbo Fu.
\newblock Sketchhairsalon: Deep sketch-based hair image synthesis.
\newblock {\em arXiv preprint arXiv:2109.07874}, 2021.

\bibitem{yang2020deep}
Shuai Yang, Zhangyang Wang, Jiaying Liu, and Zongming Guo.
\newblock Deep plastic surgery: Robust and controllable image editing with human-drawn sketches.
\newblock In {\em Computer Vision--ECCV 2020: 16th European Conference, Glasgow, UK, August 23--28, 2020, Proceedings, Part XV 16}, pages 601--617. Springer, 2020.

\bibitem{xuyao2022}
Xu~Yao, Alasdair Newson, Yann Gousseau, and Pierre Hellier.
\newblock A style-based gan encoder for high fidelity reconstruction of images and videos.
\newblock {\em European conference on computer vision}, 2022.

\bibitem{yeung2023calibrating}
Michael Yeung, Leonardo Rundo, Yang Nan, Evis Sala, Carola-Bibiane Sch{\"o}nlieb, and Guang Yang.
\newblock Calibrating the dice loss to handle neural network overconfidence for biomedical image segmentation.
\newblock {\em Journal of Digital Imaging}, 36(2):739--752, 2023.

\bibitem{Yu-ECCV-BiSeNet-2018}
Changqian Yu, Jingbo Wang, Chao Peng, Changxin Gao, Gang Yu, and Nong Sang.
\newblock Bisenet: Bilateral segmentation network for real-time semantic segmentation.
\newblock In {\em European Conference on Computer Vision}, pages 334--349. Springer, 2018.

\bibitem{Zhou_2023_CVPR}
Zhenglin Zhou, Huaxia Li, Hong Liu, Nanyang Wang, Gang Yu, and Rongrong Ji.
\newblock Star loss: Reducing semantic ambiguity in facial landmark detection.
\newblock In {\em Proceedings of the IEEE/CVF Conference on Computer Vision and Pattern Recognition (CVPR)}, pages 15475--15484, June 2023.

\bibitem{zhu2020domain}
Jiapeng Zhu, Yujun Shen, Deli Zhao, and Bolei Zhou.
\newblock In-domain gan inversion for real image editing.
\newblock In {\em European conference on computer vision}, pages 592--608. Springer, 2020.

\bibitem{zhu2021barbershop}
Peihao Zhu, Rameen Abdal, John Femiani, and Peter Wonka.
\newblock Barbershop: Gan-based image compositing using segmentation masks.
\newblock {\em arXiv preprint arXiv:2106.01505}, 2021.

\bibitem{zhu2022hairnet}
Peihao Zhu, Rameen Abdal, John Femiani, and Peter Wonka.
\newblock Hairnet: Hairstyle transfer with pose changes.
\newblock In {\em European Conference on Computer Vision}, pages 651--667. Springer, 2022.

\bibitem{zhu2020improved}
Peihao Zhu, Rameen Abdal, Yipeng Qin, John Femiani, and Peter Wonka.
\newblock Improved stylegan embedding: Where are the good latents?
\newblock {\em arXiv preprint arXiv:2012.09036}, 2020.

\bibitem{zhu2020sean}
Peihao Zhu, Rameen Abdal, Yipeng Qin, and Peter Wonka.
\newblock Sean: Image synthesis with semantic region-adaptive normalization.
\newblock In {\em Proceedings of the IEEE/CVF Conference on Computer Vision and Pattern Recognition}, pages 5104--5113, 2020.

\end{thebibliography}
}

\clearpage
\setcounter{page}{1}

\section*{Appendix}

In this appendix, we provide additional explanations, experiments, and results:

\begin{itemize}
  \item Section \ref{sec:ablation_detail}: Detailed analysis of ablation results and additional metrics.

    \item Section \ref{sec:failure_cases}: Analyzing cases of poor performance.
  
  \item Section \ref{sec:model_training}: Model architectures, learning process and hyperparameters.
  
  \item Section \ref{sec:detail_and_inference}: Additional speed measurements in different operating modes and implementation details.

    \item Section \ref{sec:full_metrics}: Complete tables with metrics.

  \item Section \ref{sec:salon_compare}: Comparison to StyleGAN-Salon~\cite{khwanmuang2023stylegan} and LOHO~\cite{saha2021loho} on the realism metric.

    \item Section \ref{sec:visual_compare_hairnet}: Comparison with HairNet~\cite{zhu2022hairnet} and HairNeRF~\cite{chang2023hairnerf} visually and in terms of features.

  \item Section \ref{sec:color_transfer}: Our attempts to find a color transfer metric.

  \item Section \ref{sec:visual_comp}: Examples of how our method works and comparisons with others, including StyleGAN-Salon.

\end{itemize}

\section{Ablation detail}\label{sec:ablation_detail}

\subsection{Color Encoder and FS space mixing}

To prove the effectiveness of our Color Encoder and the necessity of mixing F spaces, we perform 4 experiments: configurations A -- D in \cref{table:ablations}, in which we measure metrics on 1000 random triples from the CelebA-HQ dataset. 

We first compare Baseline (config A) with configurations without mixing F spaces (config C), examining the results of the metrics we see that using full F space does not give a statistically significant gain in realism. But more importantly, if we examine the visual comparison \cref{fig:Ablations}, we see that F space without mixing does not allow us to edit the hair color with our Color Encoder.

Similarly, we experiment with config B, in which we replace the Color Encoder using the Barbershop optimization process, and config D, in which we additionally remove the mixing of F spaces. From the metrics results, we can see that using such an optimization process significantly degrades the realism metrics, which proves the effectiveness of our approach. Moreover, on the visual comparison we see that config B can still edit the hair and get the desired hair color, but for config D we see the same problem where even the optimization process cannot get the desired hair color, which confirms the need to use F space mixing.

\subsection{Rotate Encoder}

Next experiment config E, we remove the Rotate Encoder from our model, essentially leaving the generation of the targeting mask as in the CtrlHair method. In terms of metrics, removing the Rotate Encoder results in reduced realism in both FID and $\text{FID}_\text{CLIP}$. Moreover, when visually comparing ablations \cref{fig:Ablations} we see 2 problems in this approach, which were described in detail in the Barbershop article: in case of a strong pose difference either vertically or horizontally, the hair mask adapts directly, which leads to severe distortions in the final image, but Rotate Encoder allows us to eliminate them effectively.

To verify that Rotate Encoder does not mess up the desired hair shape we additionally perform a 1000 image reconstruction experiment where we transfer the hair shape and color from the image itself to itself. For this, we compute the IOU metric on the hair mask of the original and the resulting Baseline configuration image and the config E image without Rotate Encoder. As a result, the use of Rotate Encoder led to a relative decrease in IOU by 2.7\%, compared to the model without the encoder. This is a very small difference compared to the improvements seen even with small pose differences, including in \cref{fig:Ablations}. 

\subsection{Pose and Shape alignment}

We also run a series of experiments in which we remove one of our encoders each and replace it with a similar optimization problem from Barbershop, configuration F-G.

Analyzing the metrics of experiments E and F, we see a serious gap in all metrics, not to mention the running time due to the optimization processes.

If we consider the config G, the metrics get much better, but by visually evaluating the results in \cref{fig:Ablations}, we see that this optimization process strongly regularizes the desired hair shape and transfers quite differently than expected. In addition, this optimization process is very long.

This also confirms the effectiveness of our alignment step for both the task of target mask generation and inpaint not only in terms of performance but also quality.

\subsection{Refinement alignment}

The last experiment config H we consider is our own HairFast model, which is Baseline (config A) with added Refinement alignment. In addition to significant gains in metrics, we can also observe its effectiveness on visual comparison. Such Post-Processing effectively fixes Color Encoder problems, for example, in cases where it cannot preserve the original face hue due to the need to change the hair color as happened in the last example \cref{fig:Ablations}. It also effectively reverts identity and fine details such as piercings, makeup and others.

\section{Failure cases}\label{sec:failure_cases}

\begin{figure*}[t!]
  \centering
  \vspace{-0cm}
   \includegraphics[width=1\linewidth]{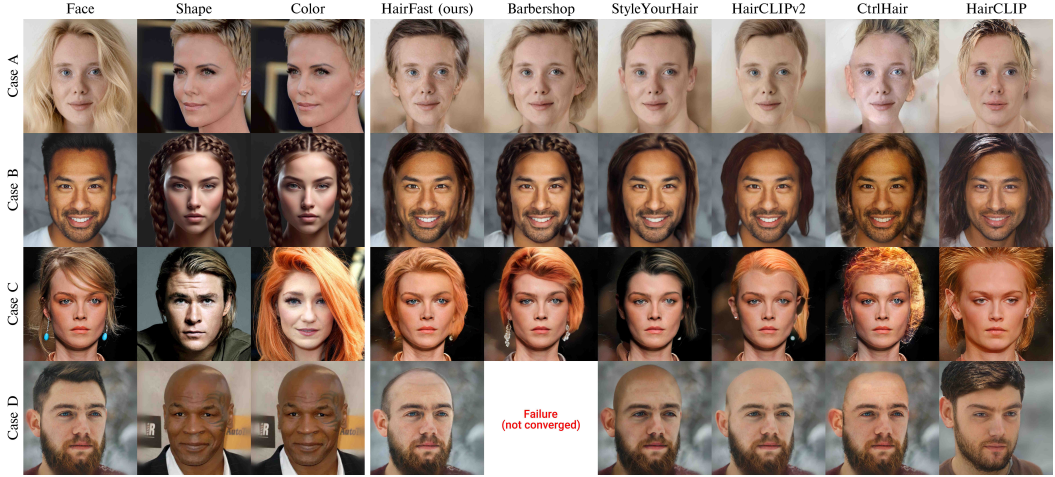}
   \caption{Failed cases. Case A poor color transfer and unrealistic inpaint. Case B unsuccessful transfer of complex texture. Case C earrings did not transfer. Case D unwanted hair in bald hairstyle transfer.}
   \label{fig:failure}
   \vspace{-0.6cm}
\end{figure*}

The \cref{fig:failure} shows examples of cases where our method does not work correctly. In particular, inpaint may not work very well for cases where long hair is replaced by short hair, in part by creating an unrealistic skin texture, or by creating shadows from past hair as in Figure \ref{fig:failure}A. In some cases, color reproduction with color encoder does not work perfectly, in particular, the problem may occur when there is a large difference in illumination, an example of failed color reproduction is shown in Figure \ref{fig:failure}A. Also, our approach does not allow to transfer hairstyles with complex textures, such as ponytails, ribbons, braids as in the figure \ref{fig:failure}B.
In addition, the model may have a problem retaining the original attributes that are exposed in the Alignment step, in which case the Refinement alignment may not be able to restore them (Figure \ref{fig:failure}C). Furthermore, our method may fail to remove some of the hair when transferring a bald hairstyle (Figure \ref{fig:failure}D). While these problems are important, they are inherent in all baseline models and we will address them in our future work.

\section{Model Training}\label{sec:model_training}

\subsection{Rotate Encoder}\label{sec:rotate_encoder}

\begin{figure*}[!ht]
  \centering
   \includegraphics[width=1\linewidth]{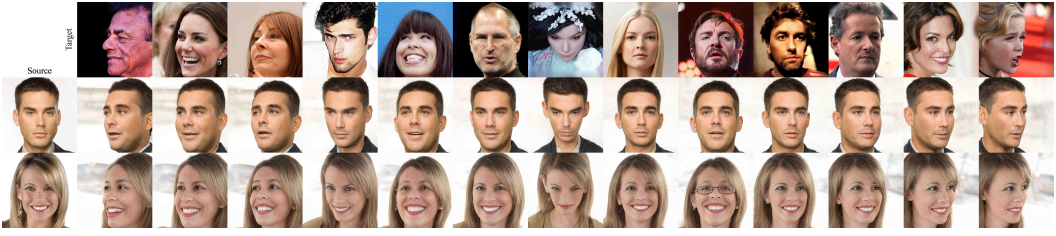}

   \caption{Demonstration of how Rotate Encoder works to rotate an image to generate the correct segmentation mask for complex pose difference images.}
   \label{fig:VisualCompRotate}
\end{figure*}

To train the Rotate Encoder, we collected 10'000 random images from FFHQ to learn how to rotate in W+ space from the E4E encoder to obtain an image with the desired pose. During the rotation, we must be able to preserve the shape of the hair.

The Rotate Encoder itself takes as input the first 6 vectors corresponding to the first StyleGAN blocks in W+ space of the source image and 6 vectors of the target image with the desired pose. And on the output it returns new 6 vectors, which should generate the image with the desired pose. We do not modify the remaining 18 - 6 vectors, which also helps us to preserve the identity in some details. An example of how Rotate Encoder works is shown in the image \cref{fig:VisualCompRotate}.

Formally, this can be described as follows:

\begin{align}
    &w_{\text{rotate}}^{1:6} = \text{Rotate}_{\text{Enc}}(w_{\text{source}}^{\text{1:6}}, w_{\text{target}}^{\text{1:6}}),\\
    &w_{\text{restore}}^{1:6} = \text{Rotate}_{\text{Enc}}(w_{\text{rotate}}^{\text{1:6}}, w_{\text{source}}^{\text{1:6}}),
\end{align}

here we additionally have $w_{\text{restore}}$, which tries to turn back the modified representation. And since we do not change the last vector representations, the following is true: $w_{\text{rotate}}^{7:18} = w_{\text{restore}}^{7:18} = w_{\text{source}}^{\text{7:18}}$. And also $w_{\text{rotate}}, w_{\text{restore}} \in \mathbb{R}^{18\times 512}$

During training, we randomly generate source image and target image pairs and train according to the following loss functions:

\begin{align}
    &H_\text{target} = E(I_{\text{target}}),\\
    &\mathcal{L}_{\text{pose}} = || H_{\text{target}} - E(G(w_{\text{rotate}})) ||_2^2.
\end{align}

In this case, $E$ -- a pre-trained model for extracting 2D face key-points, we used the STAR~\cite{Zhou_2023_CVPR} model for this purpose. For optimization, we used the first 76 key-points which corresponded to the face contour, eyebrows, eyes and nose. Thus $H_\text{target} \in \mathbb{R}^{76\times 2}$. In this way, due to $\mathcal{L}_{\text{pose}}$ loss, we train the model to rotate the image to the required pose for hair alignment.

The following loss functions are used to keep the original shape of the hair:

\begin{align}
    &\mathcal{L}_{\text{recon}} = || w_{\text{source}}^{\text{E4E}} - w_{\text{restore}} ||_2^2,\\
    &\mathcal{L}_{\text{id}} = \text{ArcFace}(I_{\text{source}}, G(w_{\text{rotate}})).
\end{align}

Here the main loss function for attribute conservation is $\mathcal{L}_{\text{recon}}$, it is what motivates the model to learn transformations that do not change the attributes of the image other than its pose. In this case, when we have rotated the image and requires to rotate it to the original pose, we know the ground truth and in this loss function we just take L2 between it. The loss function $\mathcal{L}_{\text{id}}$ is only a guide and also helps to preserve attributes a bit when rotating.

Total final loss function:

\begin{align}
    \mathcal{L} = \frac{\lambda_\text{pose} \cdot \mathcal{L}_{\text{pose}}}{\text{EMA}_{t}(\mathcal{L}_{\text{pose}})} + \frac{\lambda_\text{recon} \cdot \mathcal{L}_{\text{recon}}}{\text{EMA}_{t}(\mathcal{L}_{\text{recon}})} + \frac{\lambda_\text{id} \cdot \mathcal{L}_{\text{id}}}{\text{EMA}_{t}(\mathcal{L}_{\text{id}})}
\end{align}

The main challenge in training this model is to find the correct coefficients for the given loss functions. The model learns to adopt the correct pose very quickly and because of this, ArcFace soon begins to dominate, and with heavy overtraining, artifacts begin to form because of it. To avoid this, we first normalize the loss functions by their exponential moving average (EMA), which we compute with a factor of $t=0.02$. This allows us to maintain the correct prioritization of the loss functions throughout training.

The final model was trained with $\lambda_\text{pose}=6, \lambda_\text{recon}=2, \lambda_\text{id}=1$. The optimizer was Adam with learning rate $1\times 10^{-4}$ and weight decay $1\times 10^{-6}$. Batch size 16.

The architecture of the model is similar to the one used in Color Encoder \cref{fig:Blending}. We predict the change of $w_{\text{source}}^{\text{1:6}} \in \mathbb{R}^{6\times 512}$, and input $w_{\text{target}}^{\text{1:6}} \in \mathbb{R}^{6\times 512}$ to the modulation layer. In total, there are 5 blocks in the model form $\mathrm{Linear}(512, 512) \to \mathrm{Modulation} \to \mathrm{LeakyReLU}(0.01)$. The block diagram of the Modulation block can be seen on the \cref{fig:Blending}. The $f_\beta$ and $g_\gamma$ are $\mathrm{Linear}(512, 512) \to \mathrm{LayerNorm}(512) \to \mathrm{LeakyReLU}(0.01) \to \mathrm{Linear}(512, 512)$.

\subsection{Color Encoder}\label{sec:blending_encoder}

To train the Color Encoder, we collect about 5800 image pairs from the FFHQ dataset. For this purpose, we run our model on 3000 triples from FFHQ and save FS spaces of source, shape and color images after FS and W+ mixing, and then run Shape Module to transfer hair shape from shape image to source image and additionally run one more time to transfer hair shape from color image to source. The resulting F spaces are also saved. This allows us to create 6000 pairs for color transfer, which we additionally filter and discard images without hair, from which we can not take the desired color.

Thus each object in the training sample consists of $I_\text{source}$, $S_\text{source}$ and $F_\text{source}^{\text{align}}$ of the original image with the hair shape from the color image, and $I_\text{color}$, $S_\text{color}$ of the color image. Our goal is to modify $S_\text{source}$ to get the same hair color as $I_\text{color}$ with the given $F_\text{source}^{\text{align}}$ tensor.

The first thing we do for model training is to prepare the masks:

\begin{align}
    &H_\text{color} = (\text{BiSeNet}(I_\text{color}) = \text{hair}),\\
    &H_\text{source} = (\text{BiSeNet}(I_\text{source}) = \text{hair}),\\
    &H_\text{align} = (\text{BiSeNet}(\text{G}(F_\text{source}^{\text{align}}, S_\text{source})) = \text{hair}),\\
    &M_\text{target} = \overline{H_{\text{source}}} \cdot \overline{H_\text{align}}.
\end{align}

Now we can apply our model:

\begin{align}
    &\text{emb}_{\text{face}} = \text{CLIP}_{\text{enc}}(I_{\text{source}}\cdot M_\text{target}),\\
    &\text{emb}_{\text{hair}} = \text{CLIP}_{\text{enc}}(I_\text{color}\cdot H_{\text{color}}),\\
    &S_\text{blend} = \text{Blend}_{\text{Enc}}(S_\text{source},\ S_\text{color},\ \text{emb}_{\text{face}},\ \text{emb}_{\text{hair}}),\\
    &I_{\text{blend}} = \text{StyleGAN}(F_\text{source}^{\text{align}},\ S_{\text{blend}}),
\end{align}

And now we can apply the following loss function, which worked better than LPIPS or other combinations:
\begin{equation}
    \begin{aligned}
        &\mathcal{L}_{\text{clip}}(I_\text{1}, I_\text{2}, M_\text{1}, M_\text{2})
        =1 - \mathrm{CosSim}_\mathrm{CLIP}(I_\text{1} \cdot M_\text{1}, I_\text{2} \cdot M_\text{2}),
    \end{aligned}
\end{equation}

here $\mathrm{CosSim}_\mathrm{CLIP}$ is the cosine distance between the embedding images from the CLIP model. The images are before that multiplied by the corresponding masks M.

We get the following loss function from here, which we optimize during training for our model.

\begin{align}
    \mathcal{L}_\text{color} &= \mathcal{L}_{\text{clip}}(I_\text{blend}, I_\text{color}, H_\text{align}, H_\text{color}),\\
    \mathcal{L}_\text{face} &= \mathcal{L}_{\text{clip}}(I_\text{blend}, I_\text{source}, M_\text{target}, M_\text{target}),\\
    \mathcal{L} &= \lambda_\text{color}\cdot \mathcal{L}_\text{color} + \lambda_\text{face}\cdot \mathcal{L}_\text{face}.
\end{align}

The final model was trained with $\lambda_\text{color}=1$ and $\lambda_\text{face}=1$. The optimizer was Adam with learning rate $1\times 10^{-4}$ and weight decay $1\times 10^{-6}$. Batch size 16.

The architecture of the Color Encoder is shown in the \cref{fig:Blending}. The $f_\beta$ and $g_\gamma$ are $\mathrm{Linear}(1536, 1024)\to \mathrm{LayerNorm}(1024)\to \mathrm{LeakyReLU}(0.01)\to \mathrm{Linear}(1024, 512)$, for a total of 5 blocks in the model.

\subsection{Refinement alignment}\label{sec:post_processing}

To train Refinement encoder, we collect 10'000 triples from FFHQ on which we run our method without this encoder. The image $I_\text{blend}$ after Color Encoder and the original face image $I_\text{source}$ are the object of the training sample.

The training of this stage consisted of three parts: FS encoder training, fusers training, and the whole model finetuning.

\subsubsection{Feature Style Encoder}

For FS encoder training, we first obtain a reconstruction of the real image:

\begin{align}
    &F_{64}^{\text{source}},\ S^{\text{source}} = FS_{\text{Enc (ours)}}(I_{\text{source}}),\\
    &F_\text{style}^{\text{source}} = \text{G}_{8}(S^{\text{source}}),\\
    &F_{64}^{\text{recon}} = \alpha\cdot F_{64}^{\text{source}} + (1-\alpha)\cdot F_\text{style}^{\text{source}},\\
    &I_\text{style} = \text{StyleGAN}(F_\text{style}^{\text{source}},\ S^{\text{source}}),\\
    &I_\text{recon} = \text{StyleGAN}(F_{64}^{\text{recon}},\ S^{\text{source}}).
\end{align}

Besides the reconstruction image from FS space here we also get the F tensor from S space and the image generated by S. Also here $\alpha$ is a parameter which is 0 at the beginning and gradually increases to 1 during the learning process. And after that we use the following loss functions:

\begin{align}
    &\mathcal{L}_{\text{id}}(I_\text{1}, I_\text{2}) = \text{ArcFace}(I_{\text{1}}, I_\text{2}),\\
    &\mathcal{L}_\text{m\_LPIPS}(I_\text{1}, I_\text{2}) = \text{m\_LPIPS}(I_\text{1}, I_\text{2}),\\
    &\mathcal{L}_{\text{recon. feat}}(F_\text{1}, F_\text{2}) = ||F_\text{1} - F_\text{2}||_2^2,
\end{align}

here $\mathcal{L}_\text{m\_LPIPS}$ is a multi-scale perceptual loss~\cite{xuyao2022} presented by the authors of FS encoder, which in addition with $\mathcal{L}_{\text{id}}$ reconstructs the original image. In turn, it will $\mathcal{L}_{\text{recon. feat}}$ help in the initial stages to obtain an F tensor similar to the F tensor created from S space, so here we throw gradients only through one F tensor, not pull them together.

Thus the final loss function for training our FS encoder:

\begin{equation}
    \begin{aligned}
    \mathcal{L} &= \lambda_\text{id} \cdot (\mathcal{L}_{\text{id}}(I_\text{source}, I_\text{style}) +\mathcal{L}_{\text{id}}(I_\text{source}, I_\text{recon}))\ +\\
    &+\lambda_\text{m\_LPIPS}\cdot (\mathcal{L}_\text{m\_LPIPS}(I_\text{source}, I_\text{style})\ +\\
    &+\mathcal{L}_\text{m\_LPIPS}(I_\text{source}, I_\text{recon}))\ +\\
    &+\lambda_\text{recon. feat}\cdot \mathcal{L}_{\text{recon. feat}}(F_\text{style}^\text{source}, F_{64}^{\text{source}})\ +\\
    &+\lambda_\text{ADV}\cdot \mathcal{L}_\text{ADV}(I_\text{recon}),
    \end{aligned}
\end{equation}

here we also use adversarial loss function, as a discriminator we take the pre-trained one from StyleGAN and train it in the process using the classical loss function and R1 regularization.

The following parameters were used for training: $\lambda_\text{id}=0.1, \lambda_\text{m\_LPIPS}=0.8, \lambda_\text{recon. feat}=0.01$ and $\lambda_\text{ADV}=0.2$. Adam was used for encoder optimization with learning rate $2\times 10^{-4}$ and weight decay 0, and for discriminator learning rate $1\times 10^{-4}$, betas=(0.9, 0.999) and weight decay 0. The model was trained with a batch size 16.

\subsubsection{Fusing Encoders}

\begin{figure}[!t!]
  \centering
   \includegraphics[width=1\linewidth]{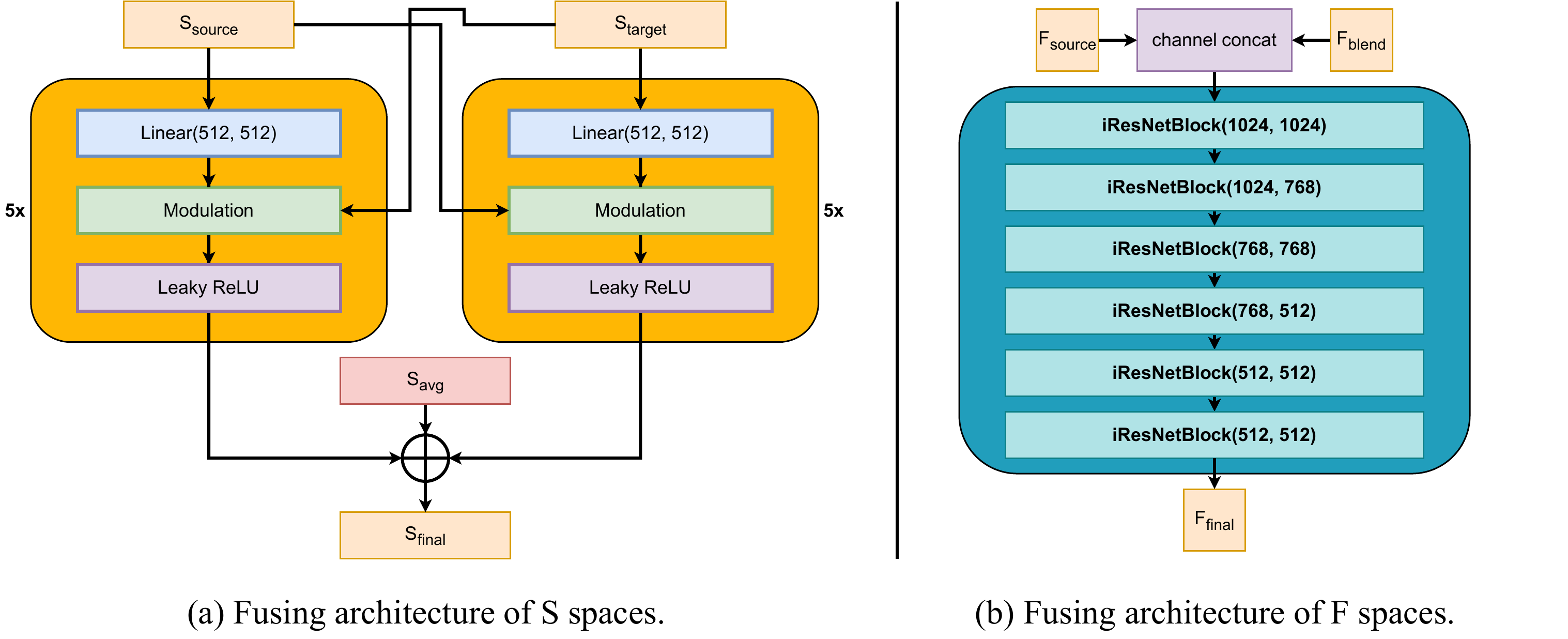}

   \caption{Diagram of fusing encoders in post processing. (a) The modulation architecture is shown in the \cref{fig:Blending}. $f_\beta$ and $g_\gamma$ in modulation are $\mathrm{Linear}(512, 512) \to \mathrm{LayerNorm}(512) \to \mathrm{LeakyReLU}(0.01) \to \mathrm{Linear}(512, 512)$. $S_\text{avg}$ is the average latent vector. (b) The architecture consists of the usual IResNet blocks.}
   \label{fig:ArchSFusing}
\end{figure}

Once we have trained the FS encoder we can proceed to train models for fusing spaces. First of all we need to get FS representations from images:

\begin{align}
    F_{64}^{\text{blend}},\ S^{\text{blend}} &= FS_{\text{Enc (ours)}}(I_{\text{blend}}),\\
    F_{64}^{\text{source}},\ S^{\text{source}} &= FS_{\text{Enc (ours)}}(I_{\text{source}}).
\end{align}

Now we can apply our models to fusing:

\begin{align}
    &F_{\text{final}} = \text{Fused}_{\text{F Enc}}(F_{64}^{\text{blend}},\ F_{64}^{\text{source}}),\\
    &S_{\text{final}} = \text{latent}_{\text{avg}} + \text{Fused}_{\text{S Enc}}(S^{\text{blend}},\ S^{\text{source}}),\\
    &F_\text{style} = \text{G}_{8}(S^{\text{final}}),\\
    &I_\text{style} = \text{StyleGAN}(F_\text{style},\ S_{\text{final}}),\\
    &I_{\text{final}} = \text{StyleGAN}(F_{64}^{\text{final}},\ S_{\text{final}}).
\end{align}

Similar loss functions are used to train the encoders, but in addition we use DSC++~\cite{yeung2023calibrating} to make the segmentation mask match the original one. We also add a loss function for inpaint. First of all, we will need new masks:

\begin{align}
    &H_\text{source} = (\text{BiSeNet}(I_\text{source}) = \text{hair}),\\
    &H_\text{blend} = (\text{BiSeNet}(I_\text{blend}) = \text{hair}),\\
    &M_\text{target} = \overline{H_{\text{source}}} \cdot \overline{H_\text{blend}},\\
    &M_\text{inpaint} = \overline{M_{\text{target}}} \cdot \overline{H_\text{blend}}.
\end{align}

The inpaint is quite sensitive to the boundaries of the mask on which it is applied, so we use a special soft dilation that runs this mask through a convolution with a kernel consisting of ones in the shape of a circle of radius 25. After that we raise the result to degree 1/4, which allows us to get more concentrated values. We also create a hard mask for the $I_\text{style}$ guide as we don't want to look at the hair of this image:

\begin{align}
    &M_\text{smooth} = \mathrm{Dilation}(M_\text{inpaint}),\\
    &M_\text{hard} = M_\text{smooth} \cdot \overline{H_\text{blend}}.
\end{align}

The discriminator is the main contributor to inpaint, but to help ease the load on it we guide it with $\mathcal{L}_\text{m\_LPIPS}$ over the inpaint area using two images: $I_\text{blend}$ and $I_\text{style}$. The first one is used to guide and preserve better details, while the second image has a more correct shading.

The final loss function is as follows:

\begin{equation}
    \begin{aligned}
    \mathcal{L} &= \lambda_\text{id} \cdot (\mathcal{L}_{\text{id}}(I_\text{source}\cdot M_\text{target}, I_\text{style}\cdot M_\text{target}) +\ \\
    &+\mathcal{L}_{\text{id}}(I_\text{source}\cdot M_\text{target}, I_\text{final}\cdot M_\text{target}))\ +\\
    &+\lambda_\text{m\_LPIPS}\cdot (\mathcal{L}_\text{m\_LPIPS}(I_\text{source}\cdot M_\text{target}, I_\text{style}\cdot M_\text{target})\ +\\
    &+\mathcal{L}_\text{m\_LPIPS}(I_\text{source}\cdot M_\text{target}, I_\text{final}\cdot M_\text{target}))\ +\\
    &+\lambda_\text{m\_LPIPS}\cdot (\mathcal{L}_\text{m\_LPIPS}(I_\text{target}\cdot H_\text{blend}, I_\text{style}\cdot H_\text{blend})\ +\\
    &+\mathcal{L}_\text{m\_LPIPS}(I_\text{target}\cdot H_\text{blend}, I_\text{final}\cdot H_\text{blend}))\ +\\
    &+\lambda_\text{recon. feat}\cdot \mathcal{L}_{\text{recon. feat}}(F_\text{style}, F_{\text{final}})\ +\\
    &+\lambda_\text{DSC++}\cdot \mathcal{L}_{\text{DSC++}}(\text{BiSeNet}(I_\text{blend}), \text{BiSeNet}(I_\text{final}))\ +\\
    &+\lambda_\text{inpaint}\cdot (\mathcal{L}_\text{m\_LPIPS}(I_\text{style}\cdot M_\text{hard}, I_\text{final}\cdot M_\text{hard})\ +\\
    &+\mathcal{L}_\text{m\_LPIPS}(I_\text{blend}\cdot M_\text{smooth}, I_\text{final}\cdot M_\text{smooth}))\ +\\
    &+\lambda_\text{ADV}\cdot \mathcal{L}_\text{ADV}(I_\text{final}).
    \end{aligned}
\end{equation}

The following parameters were used for training: $\lambda_\text{id}=0.1, \lambda_\text{m\_LPIPS}=0.4, \lambda_\text{recon. feat}=0.01, \lambda_\text{DSC++}=0.1, \lambda_\text{inpaint}=0.2$ and $\lambda_\text{ADV}=0.2$. Adam was used for encoders optimization with learning rate $2\times 10^{-4}$ and weight decay 0, and for discriminator learning rate $3\times 10^{-4}$, betas=(0.9, 0.999) and weight decay 0. The model was trained with a batch size 16.

After we have trained the fusers, we unfreeze the FS encoder and retrain the whole model with the same parameters, but with half the learning rate.

The fusing architecture of S and F spaces is presented in \cref{fig:ArchSFusing}.

\section{Operating modes}\label{sec:detail_and_inference}

Each module of the method has its own strict function, but not all modes of operation require all modules.

\subsection{Transferring the desired color}

Thus, for the task of changing only the hair color, we do not need to run the Alignment module, since we do not need to change the hair shape.

In this mode, our method performs in 0.52 and 0.27 seconds on V100 and A100, respectively, while HairCLIP performs in 0.31 and 0.25 seconds. This enables us to achieve nearly identical speed as the fastest hair transfer methods, but still have twice the quality according to the realism metrics.

\subsection{Transfer of the desired shape}

Similarly, for the task of transferring only the hair shape, we do not need to run the Pose alignment module for the Color Encoder, since the hair shape of the color matches the original hair shape that can be obtained from BiSeNet. However, the Color Module must still be run for this mode, even though we want to keep the original hair color. This is necessary due to the fact that when we transfer the hair shape, the hair color from the shape image may leak into the F space, similar to what is seen in the experiments without mixing F spaces in the ablation \cref{fig:Ablations}. In addition, Color Encoder recovers lost details during the inversion stage, which also improves the quality.

In this mode, our method performs in 0.71 and 0.40 seconds on V100 and A100, respectively. While in both shape and color transfer mode, the run times were 0.78 and 0.52 seconds.

\subsection{Transferring hairstyle and color from one image.}

This mode as well as the previous one formally works with only 2 images, which reduces the load on the encoders and also we do not need to re-run the Pose alignment module for Color Encoder.

Like the previous one, it performs in 0.71 and 0.40 seconds on V100 and A100, respectively.

\section{Complete tables}\label{sec:full_metrics}

\begin{table*}[h!]
\centering
\vspace{-0.3cm}
\caption{All metrics were measured on the same pre-selected triples of images (face, shape and color) from the CelebaHQ~\cite{karras2017progressive} dataset. Then, applying the method, FID was measured on the original dataset and the modified dataset. $\text{FID}_{\text{CLIP}}$~\cite{Kynkaanniemi2022} was counted similarly to FID using the torchmetrics library, but a CLIP encoder was used instead of Inception V3. Running time was measured as the median time among a bunch of method runs, without taking into account models initialization and loading/saving images to disk.}
\resizebox{\textwidth}{!}{
\begin{tabular}{lcccccccccc}
\toprule
\multirow{2}{*}{\textbf{Model}}         & \multicolumn{4}{c}{FID↓} & \multicolumn{4}{c}{$\text{FID}_{\text{CLIP}}$↓} & \multicolumn{2}{c}{Time (s)↓} \\
\cmidrule(lr){2-5}
\cmidrule(lr){6-9}
\cmidrule(lr){10-11}
     & full & both & color & shape & full & both & color & shape & A100   & V100 \\ \midrule
HairCLIP~\cite{wei2022hairclip}	& 34.95	& 40.68	& 40.08	& 42.92	& 12.20	& 13.32 & 10.94	& 13.44 & \boldmath$0.28$  & \boldmath$0.36$ \\
HairCLIPv2~\cite{wei2023hairclipv2} & $\underline{14.28}$ & $\underline{23.37}$ & 20.21 & $\underline{23.90}$ & 10.98 & 12.14 & 6.55 & 10.06 & 112 & 221 \\
CtrlHair~\cite{guo2022gan}	& $15.10$	& 24.81	& \boldmath$19.65$	& 25.60	& 9.52	& 10.42	& $\underline{3.62}$	& 9.59 & 6.57 & 7.87 \\
StyleYourHair~\cite{kim2022style}	& -	& 25.90	& -	& -	& -	& 10.91	& -	& - & 84 & 239 \\
Barbershop~\cite{zhu2021barbershop}	& 15.94	& 24.52	& 20.54 & 24.08	& $\underline{7.07}$	& $\underline{8.12}$	& 3.89	& $\underline{6.76}$ & 213 & 645 \\
HairFast (ours) & \boldmath$13.12$	& \boldmath$22.71$	& $\underline{20.17}$	& \boldmath$23.36$ & \boldmath$5.12$	& \boldmath$6.06$	& \boldmath$3.00$	& \boldmath$5.34$ & $\underline{0.41}$ & $\underline{0.78}$ \\
\bottomrule
\end{tabular}
}
\label{table:table_FID_final_full}
\vspace{-0.1cm}
\end{table*}

Analyzing the complete \cref{table:table_FID_final_full}, our method performs better on all cases according to the $\text{FID}_{\text{CLIP}}$ metric. But we lose to CtrlHair when transferring only hair color by FID metric because in this case CtrlHair blends almost the whole image except hair, which is strongly encouraged by the metric.

\begin{table}[h!]
\vspace{-0.0cm}
\caption{\textbf{Pose Metrics}. For this metrics, we consider color and shape transfer from the target image to the source image. For all pairs of images we calculate MAE of key facial points and split the pairs into three equal folds, which correspond to weak, medium and high pose-difference on which the corresponding metrics were measured. There were 1000 image pairs taken from CelebA-HQ. \textbf{Reconstruction Metrics}. For this each method is started on the task of transferring the color and shape of the hairstyle from itself to itself, thus at the end we measure the metrics with the original image. There were 3000 images taken from CelebA-HQ.}
\resizebox{\textwidth}{!}{
    \begin{tabular}{l*{8}c}
    \toprule
    \multirow{3}{*}{\textbf{Model}} & \multicolumn{4}{c}{\textbf{Pose metrics}} & \multicolumn{4}{c}{\textbf{Reconstruction metrics}} \\
    \cmidrule(lr){2-5} \cmidrule(lr){6-9}
    & \multicolumn{2}{c}{FID↓} & \multicolumn{2}{c}{$\text{FID}_{\text{CLIP}}$↓} \\
    \cmidrule(lr){2-3} \cmidrule(lr){4-5}
    & medium & hard & medium & hard & LPIPS↓ & PSNR↑ & FID↓ & $\text{FID}_{\text{CLIP}}$↓ \\
    \midrule
    HairCLIP~\cite{wei2022hairclip}	    & 55.77	& 54.35 & 15.53 & 15.73                                             & 0.36	& 14.08 & 35.49	& 10.48 \\
    HairCLIPv2~\cite{wei2023hairclipv2} &  44.62 & $\underline{45.28}$ & 14.56 & 18.66                              & 0.16 & 19.71 & 10.09 & 4.08 \\
    CtrlHair~\cite{guo2022gan}	        & 46.45	& 50.12 & 12.96 & 16.42                                             & 0.15	& 19.96	& \boldmath$8.03$	& $\underline{1.25}$ \\
    StyleYourHair~\cite{kim2022style}	& 46.32	& 47.19 & 13.70 & 15.93                                             & 0.14	& $\underline{21.74}$	& 10.69	& 2.73 \\
    Barbershop~\cite{zhu2021barbershop}	& $\underline{44.08}$	& 46.13 & $\underline{11.27}$ & $\underline{13.30}$ & $\underline{0.11}$	& 21.18	& 13.73	& 2.61 \\
    HairFast (ours) &  \boldmath$43.25$ & \boldmath$44.85$ & \boldmath$8.90$ & \boldmath$10.33$                     & \boldmath$0.08$ & \boldmath$23.45$ & $\underline{9.72}$ & \boldmath$0.97$\\ 
    \bottomrule
    \end{tabular}
}
\label{table:pose_diff_full}
\label{table:reconstruct_full}
\end{table}

The \cref{table:pose_diff_full} shows the full reconstruction results, we only lose to CtrlHair on the FID metric due to Poisson mixing, which is strongly encouraged by the metric.

\section{Realism after editing by StyleGAN-salon}\label{sec:salon_compare}

\begin{table}[ht!]
\centering
\small
\caption{The task of transferring hair color and shape from the target image to the original image. Selected 450 pairs from the FFHQ dataset by the authors of StyleGAN-Salon. Images of method results are provided by the authors of StyleGAN-Salon, we compute the images for CtrlHair and our method. The table shows the realism metrics as well as the RMSE of key points of the generated face and the original face, also as this metric is shown by the authors of StyleGAN-Salon.}
\begin{tabular}{lcccc}
\toprule
\textbf{Model} & Based & FID↓ & $\text{FID}_{\text{CLIP}}$↓ & RMSE↓ \\
\midrule
CtrlHair~\cite{guo2022gan}	   & Enc. & 52.14	& 15.27 & 5.48 \\
LOHO~\cite{saha2021loho}	       & Opt. & 50.24	& 15.12 & 8.16 \\
HairCLIPv2~\cite{wei2023hairclipv2}	   & Opt. & $\underline{48.95}$ & 14.59 & 5.66 \\
StyleYourHair~\cite{kim2022style}  & Opt. & 52.10	& 12.68 & 7.14 \\
Barbershop~\cite{zhu2021barbershop}	   & Opt. & 50.76	& 10.37 & 7.31 \\
StyleGAN-Salon~\cite{khwanmuang2023stylegan} & Opt. & 50.32	& $\underline{9.91}$ & $\underline{4.65}$ \\
HairFast (ours) & Enc. & \boldmath$47.15$	& \boldmath$8.45$ & \boldmath$4.52$\\
\bottomrule
\end{tabular}
\label{table:stylegan-salon}
\end{table}

The authors of the StyleGAN-Salon~\cite{khwanmuang2023stylegan} method have not yet published their code, which makes them hard to compare, but they do provide a small dataset with an inference of many basic models, including LOHO~\cite{saha2021loho}, one of the first optimization-based methods, as well as their own method. The StyleGAN-Salon method itself uses optimization-based methods and also PTI, which should make it very long. In the article itself, the authors talk about 21 minutes on a single input pair, but the measurements were done on a different hardware.

Unlike our past metrics, these were measured on the FFHQ dataset on which we trained our models. The authors are measured on 450 image pairs, which together did not occur in any training from our models.

Analyzing the results \cref{table:stylegan-salon} we see that we outperform all methods on all presented metrics, including the new metric RMSE measured between key points of the original face and the generated face. This metric confirms the effectiveness of our Shape alignment approach, while other methods may corrupt the shape of the original face. Also, these results confirm the effectiveness of our Refinement alignment stage, which outperforms even PTI in terms of realism.

Also see a visual comparison of our model with StyleGAN-Salon in \cref{fig:VisualCompSalon}.

\section{Comparison with HairNet and HairNeRF}\label{sec:visual_compare_hairnet}

The authors of HairNet~\cite{zhu2022hairnet} and HairNeRF~\cite{chang2023hairnerf} have not published their code at this time, nor do they provide any datasets on which to compare with them. So instead we compare with them based on features that can be seen partially in the \cref{table:table_characterization} and also visually in the images from their paper.

\subsection{Comparison with HairNet}

\begin{figure*}
  \centering
  \vspace{-0.3cm}
   \includegraphics[width=1\linewidth]{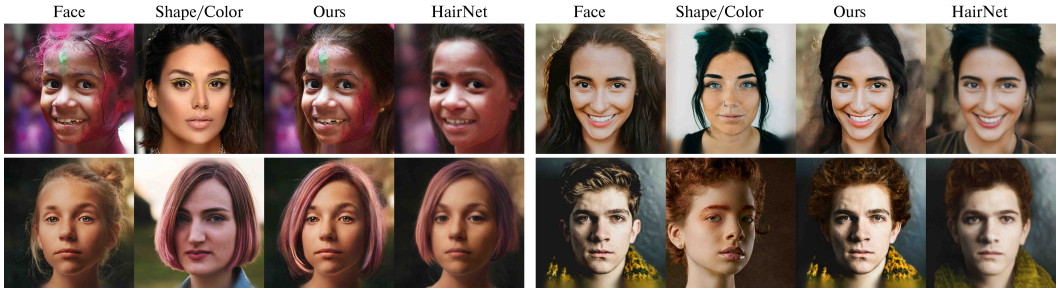}
   \caption{Visual comparison of our method with HairNet~\cite{zhu2022hairnet}.}
   \vspace{-0.3cm}
   \label{fig:hairnet}
\end{figure*}

By analyzing HairNet we can highlight some key points compared to our work: (1) HairNet uses optimization based inversion and PTI for pre-processing, which implies that it must run hundreds of times slower than our method. (2) HairNet has worse FID than Barbershop and visually worse than StyleGAN Salon. Consequently, we should expect better performance. (3) HairNet works only in low FS space, so the method retains much less details than our approach. (4) Unlike our approach, HairNet is incapable of transferring hair color independent of hair shape.

Visual analysis with images reported in HairNet paper is shown in Figure~\ref{fig:hairnet}. HairNet compared to our method tend to be poor at preserving the details of the original image and may change the identity, also the method may be worse at transferring hair texture and color.

Summarizing the above, we can say that HairNet is inferior to ours in many characteristics.

\subsection{Comparison with HairNeRF}

\begin{figure*}
  \centering
  \vspace{-0.1cm}
   \includegraphics[width=1\linewidth]{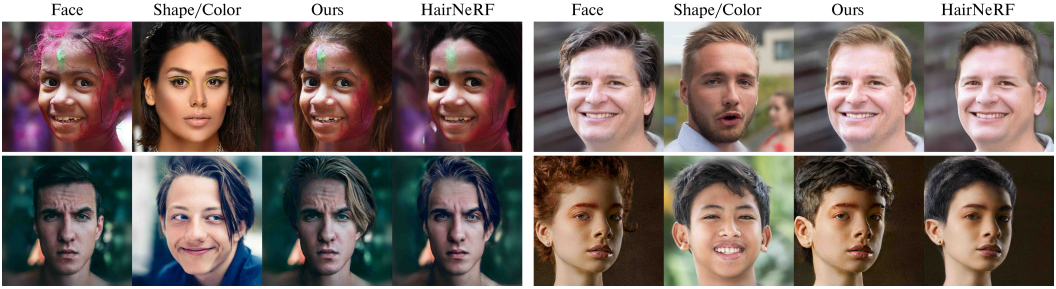}
   \caption{Visual comparison of our method with HairNeRF~\cite{chang2023hairnerf}.}
   \vspace{-0.3cm}
   \label{fig:hairnerf}
\end{figure*}

Unlike most other approaches HairNeRF~\cite{chang2023hairnerf} works on StyleNeRF~\cite{gu2021stylenerf} rather than StyleGAN, which makes it more difficult to compare features without their code. But in their work they use optimizations including image inversion with PTI, alignment and blending, which means they have to work hundreds of times longer than our approach. In addition, HairNeRF cannot independently transfer hair color or hair shape.

A visual comparison \cref{fig:hairnerf} shows the effectiveness of StyleNeRF for image alignment compared to other baseline models, but still comparable to our approach. Although HairNeRF uses PTI, our method still preserves more details of the original image and preserves the identity better.

\section{Color transfer metric}\label{sec:color_transfer}

In our work, we also try to find a metric for the quality of hair color transfer, other methods have not provided anything similar before. Our attempts to find a metric were aimed at using different hair loss functions or estimation models that were applied on the results when transferring random color and desired color. But no one model found a statistical difference between these results, which did not allow us to make a conclusion about the quality of the color transfer.

But we were able to get one of the estimates that showed statistical significance between random and set experiments. To do this, we convert the color image and the final image into HSV format, from which we take the pixels corresponding to the eroded hair mask for each image. After that, for each pixel coordinate corresponding to hue, saturation and value we construct a discrete distribution of values over 500 bins. Finally, we compute the similarity of the resulting discrete distributions using Jensen-Shannon divergence. The average results for 1000 random experiments from CelebA-HQ are summarized in Table \cref{table:color-metric}.

\begin{table}
\centering
\vspace{-0.1cm}
\caption{A color transfer metric that measures the Jensen-Shannon divergence between the histogram of the hue, saturation and value channel distributions of the hair of resulting image and the target hair image in HSV format. For this metric, we consider the color and shape transfer from the target image to the source image. 1000 image pairs from CelebA-HQ are taken.}
\small
\begin{tabular}{lccc}
\toprule
\textbf{Model} & H mean↓ & S mean↓ & V mean↓ \\
\midrule
HairCLIP~\cite{wei2022hairclip}	  & \boldmath$0.346$	& 0.275 & 0.265 \\
HairCLIPv2~\cite{wei2023hairclipv2}	  & 0.418 & 0.291 & 0.278 \\
CtrlHair~\cite{guo2022gan}	  & $\underline{0.349}$	& \boldmath$0.249$ & \boldmath$0.236$ \\
StyleYourHair~\cite{kim2022style} & 0.429	& 0.310 & 0.290 \\
Barbershop~\cite{zhu2021barbershop}	  & 0.405	& 0.266 & $\underline{0.249}$ \\
HairFast (ours) & 0.357 & $\underline{0.262}$ & 0.269 \\
\bottomrule
\end{tabular}
\label{table:color-metric}
\end{table}

Unfortunately, this metric doesn't take into account too many factors, such as things like lighting in the images, but it still allows us to draw some conclusions. 

So our method shows very good results for hue and saturation channels, which indicates quite accurate color reproduction. This correlates with the visual \cref{fig:VisualComp} and \cref{fig:VisualComp2} comparison.

\section{Visual comparison}\label{sec:visual_comp}

\begin{figure*}
  \centering
  \vspace{-0.6cm}
   \includegraphics[width=1\linewidth]{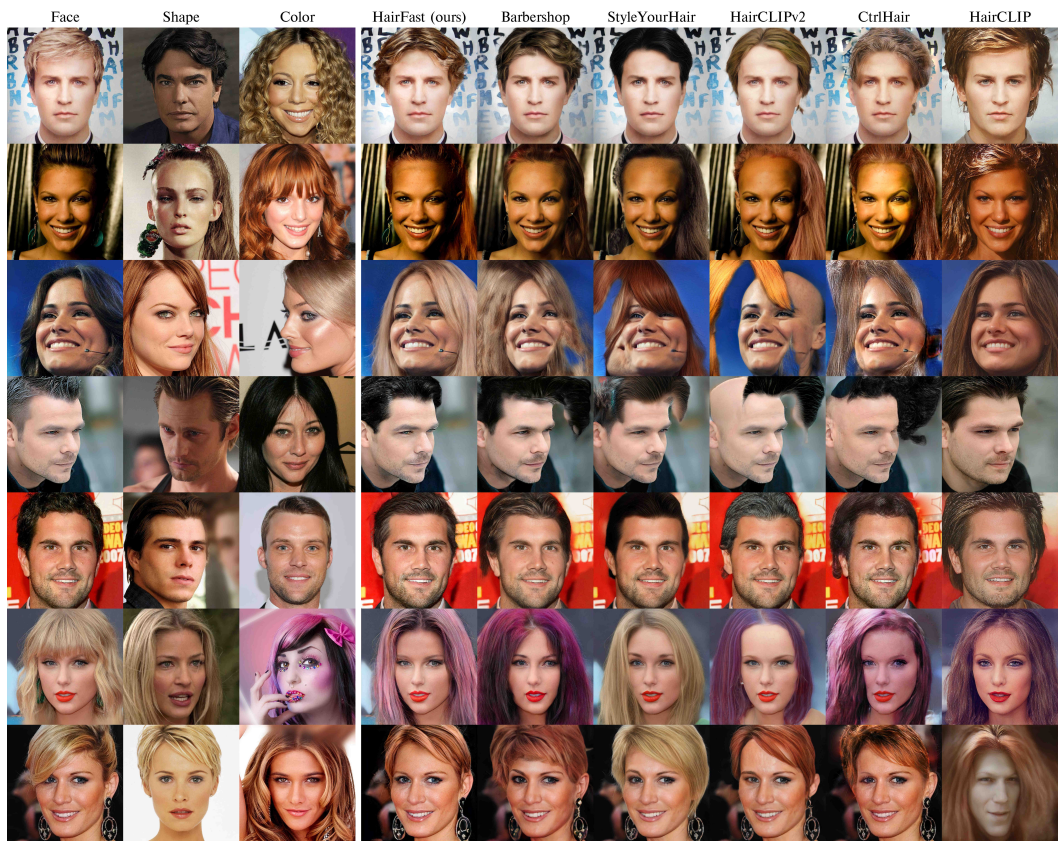}

   \caption{The first part of a visual comparison of the performance of the methods. All methods transfer hair shape and color from given images, except StyleYourHair, which due to its limitations transfers both shape and color only from Shape images.}
   \label{fig:VisualComp}
\end{figure*}

\begin{figure*}
  \centering
  \vspace{-0.6cm}
   \includegraphics[width=1\linewidth]{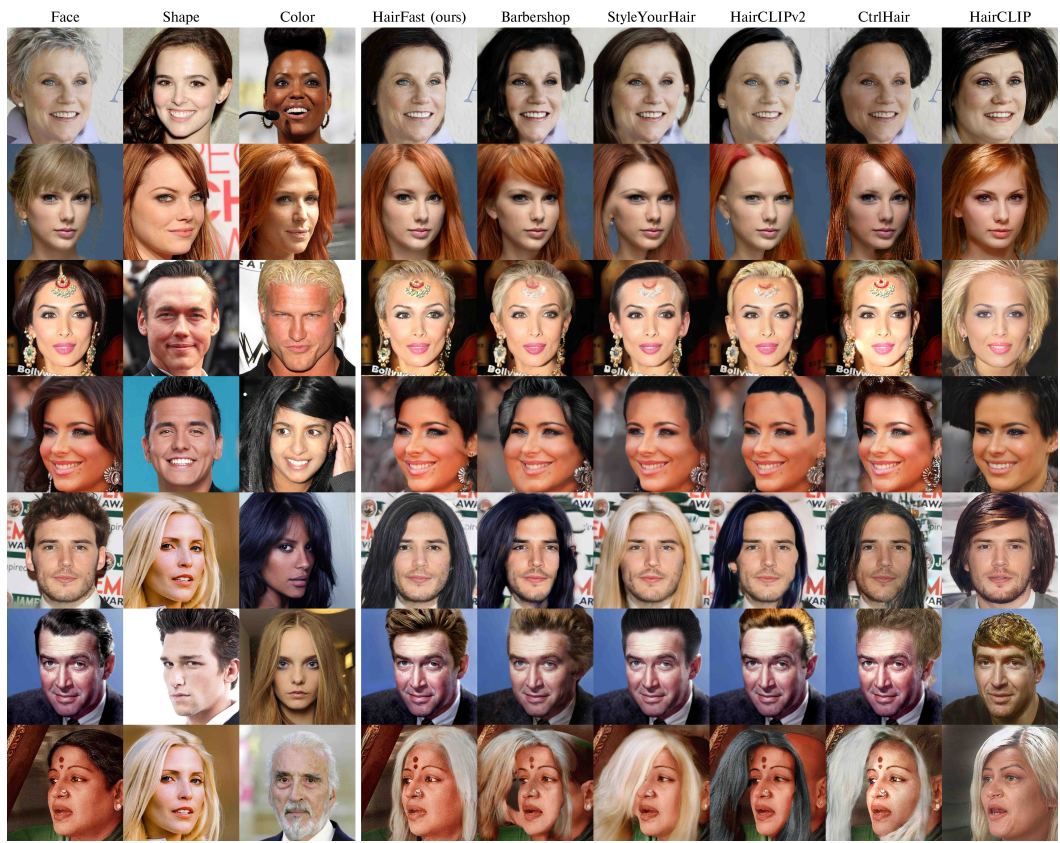}

   \caption{The second part of a visual comparison of the performance of the methods. All methods transfer hair shape and color from given images, except StyleYourHair, which due to its limitations transfers both shape and color only from Shape images.}
   \label{fig:VisualComp2}
\end{figure*}

The \cref{fig:VisualComp} and \cref{fig:VisualComp2} shows the results of a visual comparison with the Barbershop~\cite{zhu2021barbershop}, StyleYourHair~\cite{kim2022style}, HairCLIPv2~\cite{wei2023hairclipv2}, CtrlHair~\cite{guo2022gan} and HairCLIP~\cite{wei2022hairclip} methods. StyleYourHair cannot transfer the desired color from a different image, so it tried to transfer the color from the shape image. Also, for visual comparison, we disabled StyleYourHair's horizontal flip feature, which was enabled when we calculated our metrics in the main part. This feature allows StyleYourHair to check if the reflected image will have a smaller pose difference than the original image and transfer the hairstyle from it. We disabled it in this compare to see how their approach to image rotation works compared to ours, and because most hairstyles are asymmetrical and often want to be transferred as they are in the example.

Analyzing the \cref{fig:VisualComp} and \cref{fig:VisualComp2} results, in addition to the speed gain, we would like to mention the excellent quality, which in most cases outperforms other methods. For example, our method successfully captures the desired hue in all submitted images, while other methods fail. In addition, our method is much better at preserving the identity of the face, its details and its surroundings. And with all this, our method handles well the difficult cases related to pose differences, while even StyleYourHair, which is focused on this task, does worse.

\begin{figure*}
  \centering
  \vspace{-0.6cm}
   \includegraphics[width=1\linewidth]{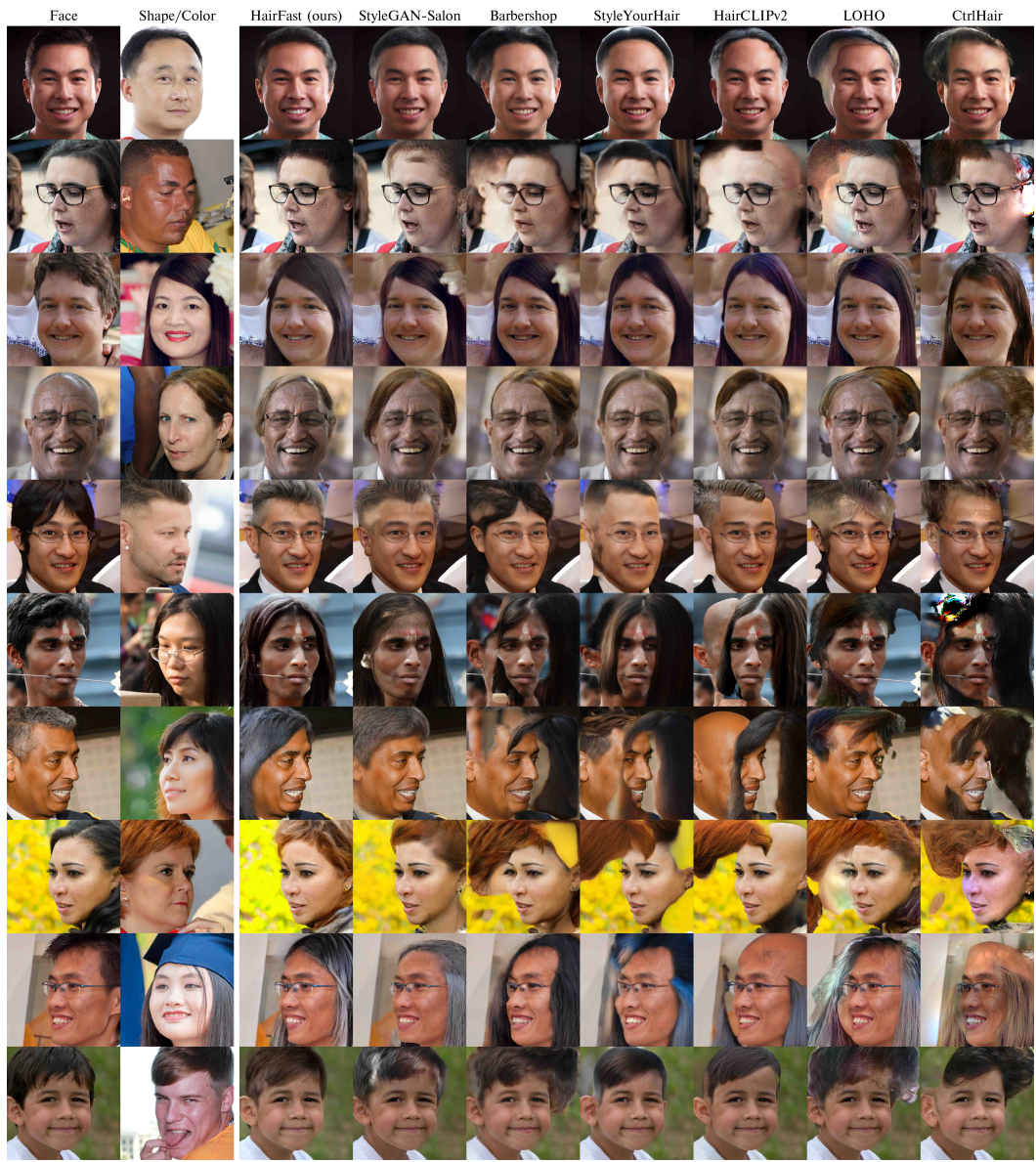}

   \caption{Visual comparison with pictures presented by the authors of StyleGAN Salon, where we are additionally compared to their model as well as LOHO.}
   \label{fig:VisualCompSalon}
\end{figure*}

We also do a visual comparison with StyleGAN-Salon in the image \cref{fig:VisualCompSalon}. Although their method produces PTI for each example to preserve more details of the original image, our large encoder approach performs much better and preserves much more details. This in particular allows us to better preserve the identity of the face and attributes such as glasses. This is also confirmed by the \cref{table:stylegan-salon} metrics. In addition, StyleGAN-Salon's 3D image rotation approach with hair shape sometimes creates artifacts on the hair, and also sometimes changes long hair to short hair because of this.

\begin{figure*}
  \centering
  \vspace{-0.3cm}
   \includegraphics[width=1\linewidth]{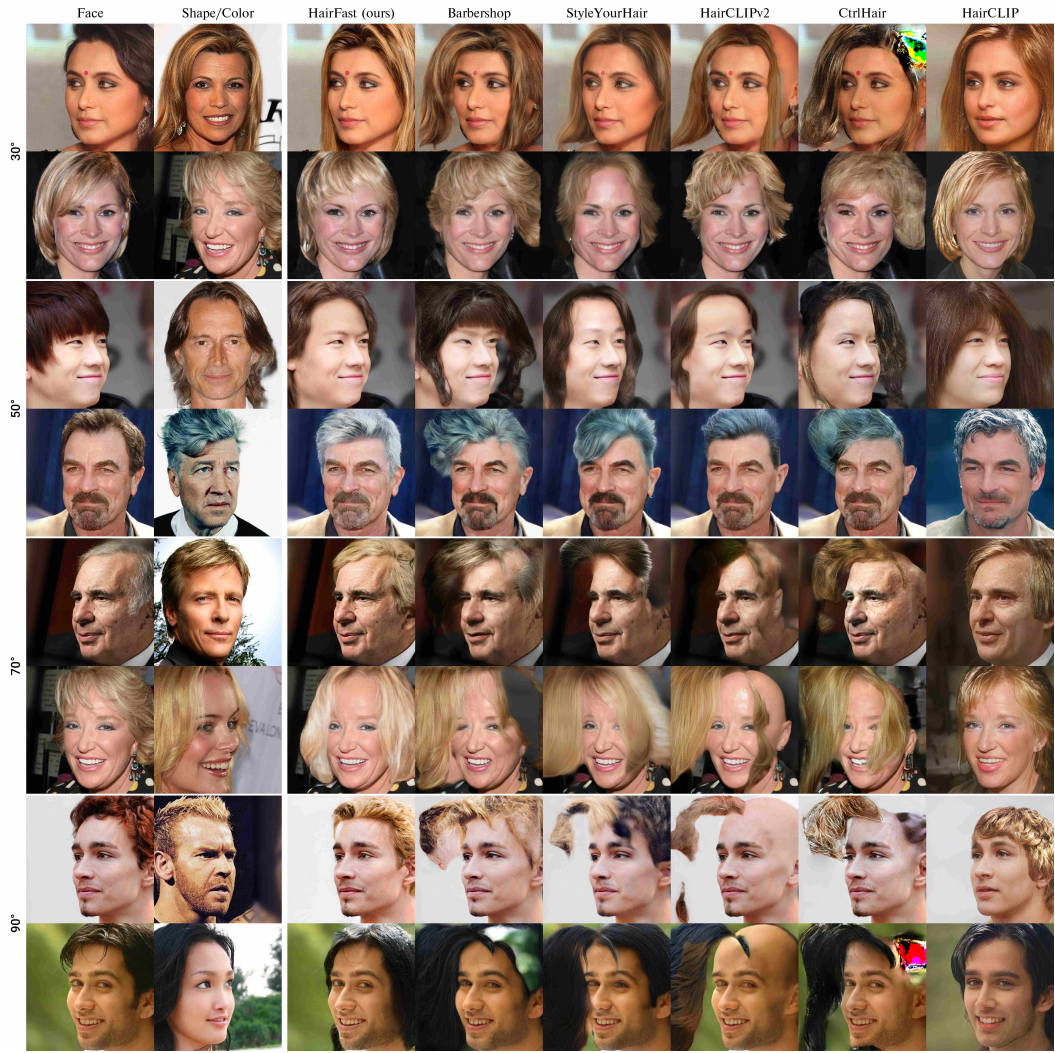}
   \caption{Detailed comparison of methods with different face rotation, the image shows cases with 30, 50, 70 and 90 degree rotation. Our method to rotation is significantly superior to other methods of hair transfer.}
   \vspace{-0.3cm}
   \label{fig:pose_diff_image}
\end{figure*}

You can also see in the \cref{fig:pose_diff_image} image how our method compares to other methods for large pose differences, in particular 30, 50, 70 and 90 degree differences. Our approach with Rotate Encoder significantly outperforms other methods, including StyleGANSalon which is focused on solving this problem. At the same time, our method can handle even the largest pose differences without causing artifacts and still keep the hairstyle recognizable.

\begin{figure*}
  \centering
  \vspace{-0.3cm}
   \includegraphics[width=1\linewidth]{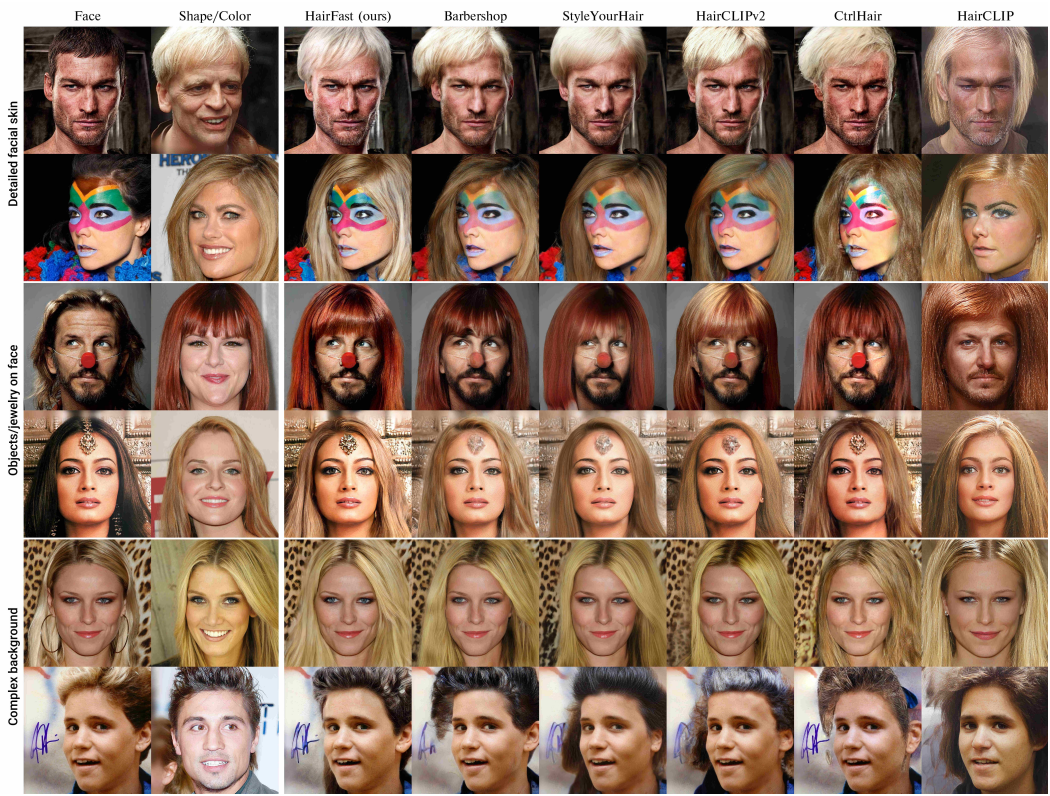}
   \caption{The image shows the comparison of our method with the others on various difficult cases for the method: complex face, objects on the face and complex background.}
   \vspace{-0.3cm}
   \label{fig:face_details_image}
\end{figure*}

The last visual experiment can be observed in the \cref{fig:face_details_image} image. Here we compare our method with others in various complex cases including complex face, details on the face and complex background. Our Refinement alignment step significantly outperforms other detail preservation approaches by producing significantly more realistic images. While CtrlHair also outperforms other methods using Poisson blending, this approach leaves noticeable artifacts at mask boundaries.

\end{document}